\documentclass[pdflatex,sn-mathphys-num]{sn-jnl}

\usepackage{graphicx}%
\usepackage{multirow}%
\usepackage{amsmath,amssymb,amsfonts}%
\usepackage{amsthm}%
\usepackage{mathrsfs}%
\usepackage[title]{appendix}%
\usepackage{xcolor}%
\usepackage{textcomp}%
\usepackage{manyfoot}%
\usepackage{booktabs}%
\usepackage{algorithm}%
\usepackage{algorithmicx}%
\usepackage{algpseudocode}%
\usepackage{listings}%
\usepackage{lmodern}
\usepackage{anyfontsize} 
\usepackage{enumerate} 
\begin{document}

\title[AI Can Enhance Creativity in Social Networks]{AI Can Enhance Creativity in Social Networks}


\author[1]{\fnm{Raiyan Abdul} \sur{Baten}}\email{rbaten@usf.edu}
\author[1]{\fnm{Ali Sarosh} \sur{Bangash}}\email{alibangash@usf.edu}
\author[1]{\fnm{Krish} \sur{Veera}}\email{krishv@usf.edu}

\author[2]{\fnm{Gourab} \sur{Ghoshal}}\email{gghoshal@pas.rochester.edu}

\author*[3]{\fnm{Ehsan} \sur{Hoque}}\email{mehoque@cs.rochester.edu}

\affil[1]{\orgdiv{Department of Computer Science and Engineering}, \orgname{University of South Florida}, \orgaddress{\street{4202 E Fowler Avenue}, \city{Tampa}, \state{FL-33620}, \country{USA}}}

\affil[2]{\orgdiv{Department of Physics and Astronomy}, \orgname{University of Rochester}, \orgaddress{\street{464 Bausch \& Lomb Hall}, \city{Rochester}, \state{NY-14627}, \country{USA}}}

\affil[3]{\orgdiv{Department of Computer Science}, \orgname{University of Rochester}, \orgaddress{\street{3013 Wegmans Hall}, \city{Rochester}, \state{NY-14627}, \country{USA}}}

\abstract{Can peer recommendation engines elevate people's creative performances in self-organizing social networks? Answering this question requires resolving challenges in data collection (e.g., tracing inspiration links and psycho-social attributes of nodes) and intervention design (e.g., balancing idea stimulation and redundancy in evolving information environments). We trained a model that predicts people's ideation performances using semantic and network-structural features in an online platform. Using this model, we built \texttt{SocialMuse}, which maximizes people's predicted performances to generate peer recommendations for them. We found treatment networks leveraging \texttt{SocialMuse} outperforming AI-agnostic control networks in several creativity measures. The treatment networks were more decentralized than the control, as \texttt{SocialMuse} increasingly emphasized network-structural features at large network sizes. This decentralization spreads people's inspiration sources, helping inspired ideas stand out better. Our study provides actionable insights into building intelligent systems for elevating creativity.}


\maketitle


`Lone genius' is a myth; creative ideas rarely emerge from a social vacuum. From music to scientific research, or from architecture to digital painting---our world heavily relies on people's ability to generate novel and appropriate ideas that stand out from prior ideas (``creativity'')~\cite{hofstra2020diversity,uzzi2013atypical}. In search of those ideas, people interact with and find inspiration from others all the time. Consider a graphic designer who takes inspiration from fellow designers on a portfolio website or an academic researcher who finds inspiration from other academics online. This social aspect of idea generation can be modeled using \textit{self-organizing social networks}, where the nodes denote idea-generating humans, the edges connect node-pairs when one node takes idea inspiration from the other, and the social network structures among the idea-generators evolve temporally. Can AI intervene in the connection patterns of such social networks to elevate the creative idea-generation performances of its members?

Recent advancements at the intersection of AI and creativity have primarily focused on using AI to generate creative products directly~\cite{li2020attribute,jia2024simulbench} or to assist humans' creative workflows as a tool~\cite{davis2013toward,jeon2021fashionq,shih2011brainstorming}. Using AI to \textit{nudge the connection patterns} in a self-organizing social network to help ideators achieve better creative performance has remained largely unaddressed. In the academic research scenario mentioned above, a Ph.D. student may be inclined to seek idea inspiration from the highly visible superstars in their domain, but an online platform may recommend the student an intelligently curated set of connections to help their ideas stand out better. Such a peer recommendation engine will need to maximize the users' creative outcomes rather than the commonly employed metrics of user engagement and user satisfaction~\cite{peters2024context,liu2019characterizing,xia2023deep}. The development of such technology can have direct practical applications in online social platforms and, at the same time, help elucidate the underlying science of creativity in social networks.

We present \texttt{SocialMuse}, an intelligent peer recommendation system that senses semantic and network structural context factors and uses supervised learning to proactively nudge social network connectivity patterns toward directions of higher creative performance. Across ten independent trials ($N=420$) in a web-based experimentation platform, we show that participants in AI-informed treatment networks significantly outperformed participants in AI-agnostic control networks by producing a larger number of distinct ideas, a larger number of non-redundant ideas, more semantically diverse ideas, and better person-level best ideas in a series of idea generation tasks.

\section*{Challenges and Our Solutions}\label{challenges}
The key reasons why explorations of AI and creativity in self-organizing social networks remained under-addressed in literature can be broadly organized into two categories: (i)~data collection and (ii)~intervention design challenges.

\textbf{Data collection challenges.} \textit{First}, it is challenging to curate datasets in the wild that contain explicit and unambiguous observations of creative network \textit{edges}, i.e., the links between social sources of creative inspiration and the corresponding recipients. People may find inspiration from human and non-human sources and may not leave observable traces even on digital platforms. The elusive nature of idea inspiration thus prohibits a granular, data-driven exploration of networked creativity. 

\textit{Second}, human social networks tend to self-organize based on psycho-social cues or attributes of the social \textit{nodes}. People generally have agency over who they interact with in their social networks. Given an objective, they can choose to make or break social ties~\cite{perc2010coevolutionary}---often in response to success, age, gender, prestige, popularity, and other cues of their social partners~\cite{kelty2023don,henrich2016secret,herrmann2007humans,boyd2011cultural}. This dynamic characteristic affords opportunities in human populations that static networks cannot: for example, dynamic or self-organizing networks can promote cooperation~\cite{rand2011dynamic,szolnoki2008making}, collective intelligence~\cite{bernstein2018intermittent,almaatouq2020adaptive}, and public speaking skills~\cite{shafipour2018buildup,baten2019upskilling}. Such self-organization increases the system's complexity and complicates data collection with psycho-social-temporal granularity.

\begin{figure}[t]
    \centering
    \includegraphics[width=1\linewidth]{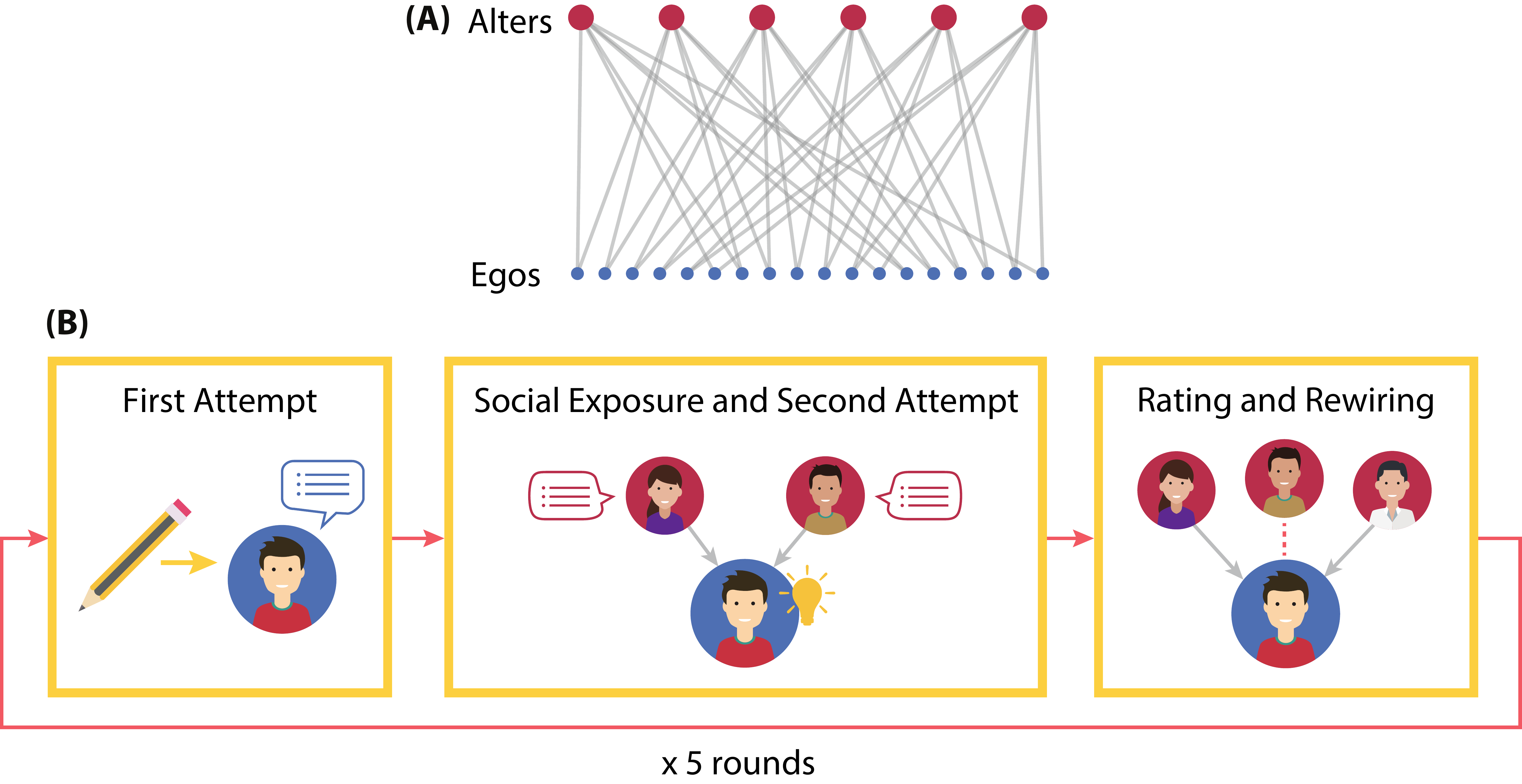}
    \caption{\textbf{Data collection and experimentation framework.} (A) The initial (round 1) network structure with 6 alter nodes and 18 ego nodes. Each ego is connected to two alters out of the six. Pre-recorded ideas of the same alters are shown to the egos of different study conditions (each with 18 egos). The entire arrangement is replicated across multiple trials with independent participants. (B) The protocol for each of the five rounds of idea generation. During attempt 1, an ego (blue node) generates ideas independently. During attempt 2, they view the ideas of the two followee alters (red nodes) and submit possible inspired ideas. Then, in the rewiring stage, the egos rate the ideas of all six alters and update which two alters to follow in the next round. }
    \label{generic_protocol}
\end{figure}

\textit{Our solution.} In prior work, we developed an online experimentation framework to address these twin data-collection challenges~\cite{baten2020creativity,baten2021cues,baten2022novel}. In the framework, we embed participants into bipartite networks of \textit{alters} (whose ideas are recorded first to be used as stimuli) and \textit{egos} (who `follow' alters of their choice for inspiration). The egos are split into different study conditions to allow systematic manipulation of desired variables (e.g., network plasticity~\cite{baten2020creativity} or psycho-social cues of the alters~\cite{baten2021cues,baten2022novel}) across conditions. The participants generate alternative usage ideas for five everyday objects (e.g., a shoe or a wooden pencil) in five consecutive rounds~\cite{guildford1978alternate}. In each round, the egos first generate ideas independently (`attempt 1'). Then, they view the ideas of two alters they are following and submit possible inspired ideas (`attempt 2'). Finally, the egos rate all the alters' ideas and optionally follow or unfollow alters to update who to take inspiration from in the next round (`rewiring'). In this bipartite network design, the egos can asynchronously complete the study since they only view the pre-recorded alters' ideas, not those of other egos. Figure~\ref{generic_protocol} summarizes the protocol. The text-based ideas are scored for creativity using manual annotation, natural language processing, and information theory-based techniques established in the literature. 

Crucially, this framework allows us to (i)~directly observe social network edges of idea inspiration (from alters to egos), (ii)~study the co-evolution patterns of the social network structures and the resulting creative performance outcomes (of the egos), and (iii)~isolate the effects of desired psycho-social variables by conducting randomized controlled experiments. 

In the current work, we build \texttt{SocialMuse} using the data and insights from our prior experiments based on this framework~\cite{baten2020creativity,baten2021cues,baten2022novel}. We design the current validation experiment using the same framework, where we manipulate the availability of AI-driven peer recommendations (i.e., alter recommendations shown to egos during the rewiring stage of each round) between study conditions.

\textbf{Intervention design challenges.} Our prior experiments show that cognitive, network-structural, and psycho-social contextual factors can influence creative performances in social networks~\cite{baten2020creativity,baten2021cues,baten2022novel}. However, designing interventions to `tweak' the contextual factors toward ideation advantage is non-trivial: 

\textit{First,} different contextual features affect creative outcomes differently in an intertwined manner, making intervention design difficult. For instance, there is a tension between semantic forces that can help one's creativity and social network structural forces that can inhibit it~\cite{baten2020creativity}. Drawing from the Associative Theory of creative cognition, external stimuli that are semantically different from one's own ideas may help a person access concepts in their long-term semantic memory that they could not on their own, potentially inspiring novel ideas in that person~\cite{kenett2023creatively,brown2002making,mednick1962associative,nijstad2006group,runco2014creativity} (please see Supplementary Text for more details). In social networks, the ideas of the most creative performers tend to have large semantic distances from commonly generated ideas~\cite{baten2020creativity}. Thus, generating peer recommendations based on idea-level semantic distances can lead to the best performers being recommended to others in the network at a disproportionately high rate.

However, a network structural counter-effect can simultaneously kick in: most people themselves tend to seek inspiration from the highest performers as they self-organize in the network~\cite{baten2020creativity}. This tendency increases the overlap in people's inspiration sources, which, in turn, increases redundancy even in their independently-generated ideas. The simple peer-recommendation heuristic of leveraging semantic distances to maximize stimulation chances can thus worsen the redundancy problem. Moreover, psycho-social contextual factors (e.g., effects of homophily and popularity cues) can also affect network connection evolution and idea redundancy patterns in social networks~\cite{baten2021cues,baten2022novel}, further complicating the intervention design problem. How an AI intervention can balance this tension between maximizing stimulation and minimizing redundancy then becomes a puzzle. Prior work suggests that \textit{partially} dispersing the degree centrality of the top ideators may help strike such a balance~\cite{baten2022novel}. Still, no guidelines exist for determining that optimal network structure, let alone achieving it.

\textit{Second,} the information environments constantly evolve as participants generate ideas and temporally rewire social ties---further exacerbating the challenge of estimating optimal network structures that the AI can seek to attain. Importantly, our AI is meant to \textit{recommend} peers rather than \textit{impose} any supposedly optimal network structure onto the participants. Thus, the intervention design must adapt to the organically changing environmental conditions during deployment.

\textit{Our solution.} We use supervised learning to learn from data how the complex interactions among multifaceted contextual factors affect creative performance outcomes. During deployment, we feed the most recent contextual information surrounding a participant to \texttt{SocialMuse} in real-time. Using the trained model, we search for the most promising social ties that maximize the participant's predicted ideation performance in the upcoming round and make peer recommendations accordingly. Thus, the peer recommendation engine strives to locally optimize each participant's ideation outcomes in a single future round, leveraging trade-offs between idea stimulation and redundancy patterns directly learned from prior data.

\section*{Overview of \texttt{SocialMuse}}\label{overview}

\begin{figure}[t]
    \centering
    \includegraphics[width=1\linewidth]{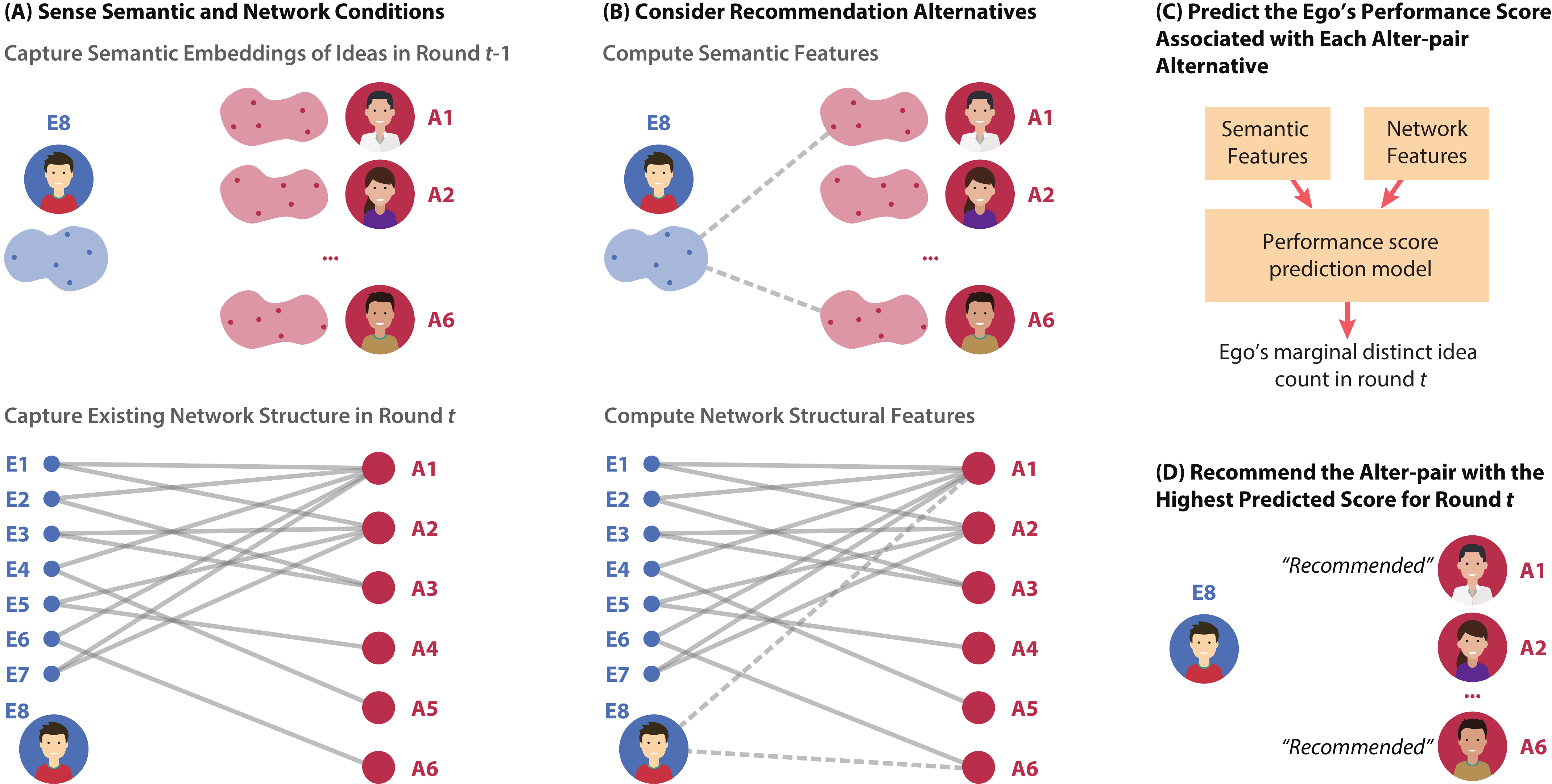}
    \caption{\textbf{Architecture of \texttt{SocialMuse}.} In this stylized example, Ego $E8$ has completed the idea-generation activities for round $t-1$. In the rewiring stage of round $t-1$, we seek to recommend ego $E8$ two alters to follow for inspiration in round $t$. Seven other egos completed round $t$ before ego $E8$. (A)~We first sense existing semantic and network-structural context factors. (B)~Using a brute-force approach, we consider all possible alter-pairs that can be recommended to $E8$ and generate semantic and network features for each alternative. The alter-pair of $\{A1, A6\}$ is being considered in the illustration, as shown with dashed lines. (C)~We use a trained model to separately predict ego $E8$'s marginal distinct idea counts in round $t$ corresponding to the features from each alter-pair alternative. (D)~We recommend the highest-scoring alter-pair to the ego.}
    \label{infograph_AI}
\end{figure}

At a high level, the \texttt{SocialMuse} system combines a (i)~Performance score prediction module with a (ii)~Recommendation generation module. The system flowchart is shown in Figure~\ref{infograph_AI}.

\textbf{Performance score prediction module.} We compile a dataset from our previous experiments to train and test a performance score prediction model (total $N_e=360$ egos, see Methods for details)~\cite{baten2020creativity,baten2021cues,baten2022novel}. We build on the simple intuition that people strive to generate ideas that stand out from prior ideas in a given task or prompt. Therefore, we take `marginal distinct idea count' as the target score in our prediction, operationalized by the number of new ideas an ego generates during attempt 2 of a given round against the pool of distinct ideas submitted by previous egos who completed that round. Each ego $e_i$ contributes four data points with target scores $s_{e_i}^t$, capturing their marginal distinct idea counts in rounds $t \in \{2, 3, 4, 5\}$. 

As input features of those data points, we consider (i)~semantic features from round $t-1$ (computed from the round $t-1$ ideas of ego $e_i$ and the actual alters they followed in round $t$) and (ii)~network structural features from round $t$ (computed from the round $t$ network comprising the alter connections of the egos upto and including $e_i$). In essence, we capture how the most recent semantic and network contextual features affect an ego's marginal distinct idea counts in a given round $t$ of idea generation. We find an XGBoost model ($\mathcal{M}$) to produce the best test-set prediction performance with $R^2=32.58\%$. The details of model training are provided in the Methods section.

\textbf{Recommendation generation module.} During deployment, at the end of round $t-1$, \texttt{SocialMuse} recommends an ego two alters (out of six) to follow for inspiration in the upcoming round $t$. Using a brute-force approach, \texttt{SocialMuse} considers all possible alter-pair alternatives and generates semantic and network features for each alternative. Using the trained model $\mathcal{M}$, \texttt{SocialMuse} then separately predicts the ego's marginal distinct idea count in attempt-2 of round $t$ corresponding to each alter-pair alternative. Finally, the system recommends the highest-scoring alter-pair to the ego along with a SHAP-value-based explanation of what the recommendation is intended to achieve~\cite{scott2017unified} (Figure~\ref{infograph_AI}; please see Methods for details).

\section*{Validation Experiment Setup}

\begin{figure}[t]
    \centering
    \includegraphics[width=0.8\linewidth]{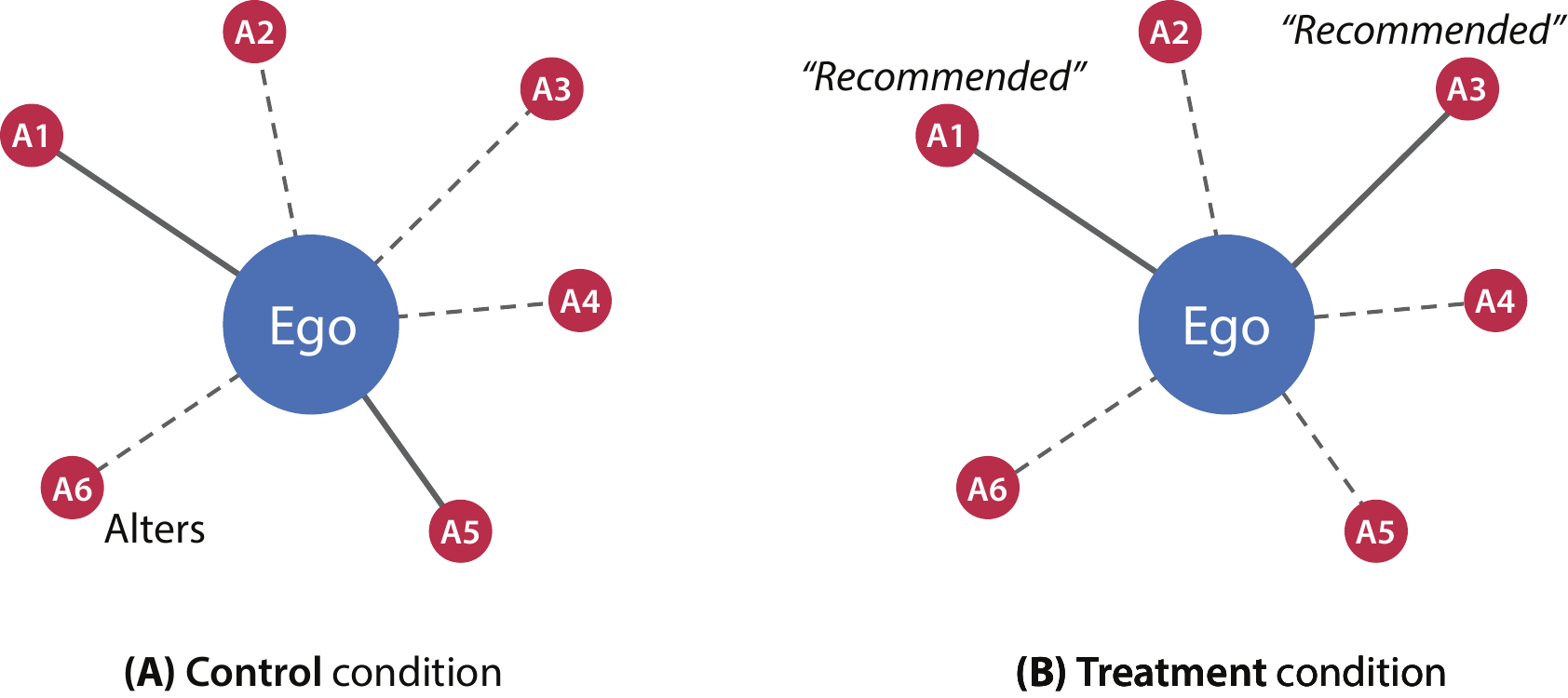}
    \caption{\textbf{Manipulating the availability of peer recommendations across conditions.} In the rewiring stage of each round, the treatment egos receive AI-generated recommendations to follow two alters out of $6$ for inspiration in the following round. The control egos do not receive any recommendations. Importantly, the egos retain the agency to pay heed to the recommendation or not. In this illustration, solid lines show the two alter connections an ego chooses to make.}
    \label{reco_ai}
\end{figure}

We conduct a randomized controlled experiment in the virtual laboratory with (i)~AI-informed treatment and (ii)~AI-agnostic control conditions. We use the same data collection framework explained in the `Challenges and Our Solutions' Section. We collect data from ten independent trials, each comprising $6$ alters, $18$ egos in the treatment condition, and $18$ egos in the control condition (total $N=420$). Egos from both conditions within a trial seek inspiration from the same set of alters, where they can view the pseudo-usernames and text-based ideas of the alters in the same order. In the treatment condition, two alters of the six are recommended to each ego during the rewiring stage of each round. The control condition has no recommendations, marking the only difference between the two conditions (Figure~\ref{reco_ai}). Thus, any difference in the connectivity dynamics and creative outcomes between the two conditions can be attributed to the availability of AI-made alter recommendations. The experimental protocol and creativity measures are elaborated in Methods.

\section*{Results}\label{results}
\subsection*{Individual and collective creativity are better in AI-driven networks}

\begin{figure}[t]
    \centering
    \includegraphics[width=1\linewidth]{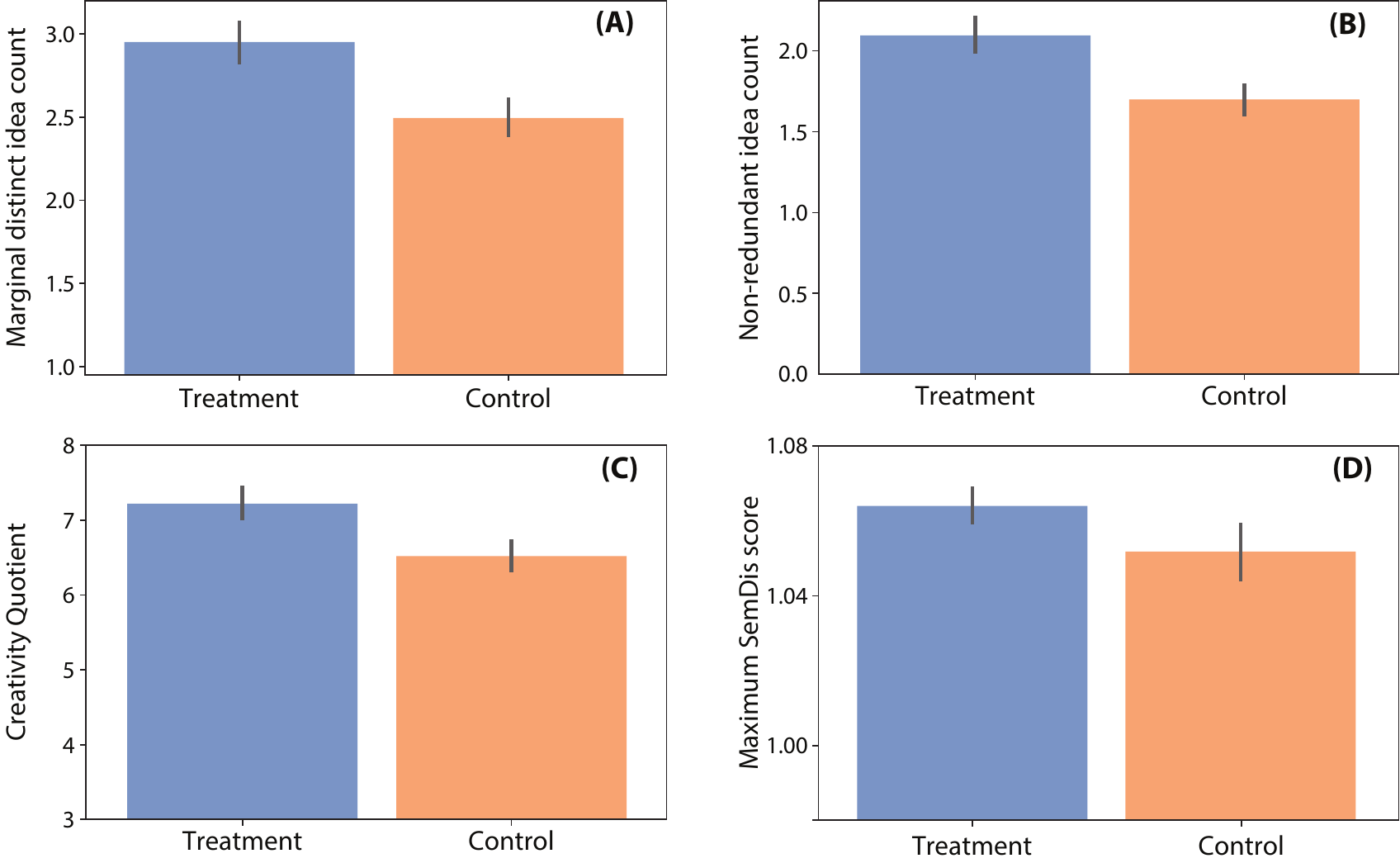}
    \caption{\textbf{Performance comparison between treatment and control conditions.} The egos in the treatment condition significantly outperform the control egos in (A) marginal distinct idea counts per round, (B) non-redundant idea counts per round, (C) Creativity Quotient per round, and (D) Maximum SemDis scores (i.e., scores of best ideas) in a given round. Whiskers denote $95\%$ C.I.}
    \label{performance_AI}
\end{figure}

The AI seeks to boost an ego's marginal distinct idea count in attempt 2 of the upcoming round against the cumulative idea pool of prior egos who completed that round (all attempts combined). Therefore, we report the egos' performances in attempt 2 in the following analyses. We find that the marginal distinct idea counts of the treatment egos (per ego per round) are indeed significantly higher than those of the control egos (Figure~\ref{performance_AI}(A); $\beta = 0.46, SE = 0.1, P<0.001$; linear mixed-effects model; Supplementary Table~\ref{marginal_nonredun_t2_art}). Furthermore, the marginal distinct idea counts show a significant downward trend as the networks grow larger (i.e., as more egos asynchronously complete the experiment; $\beta = -0.13, SE = 0.01, P<0.001$; linear mixed-effects model; Supplementary Table~\ref{marginal_nonredun_t2_lmer}). This intuitively makes sense since contributing new ideas becomes more difficult as the roundwise pool of prior ideas grows with an increasing network size. The marginal distinct idea counts do not vary significantly across rounds.

We examine the non-redundant idea counts of the egos, where an idea is defined as non-redundant if submitted by only one ego out of all the $18$ egos in the network who eventually complete the study (both attempts combined). We find that the treatment egos significantly outperform the control egos in this metric (Figure~\ref{performance_AI}(B); $\beta = 0.40, SE = 0.1, P<0.001$; linear mixed-effects model; Supplementary Table~\ref{nonredun_t2}). In other words, while the AI focuses on optimizing an ego's \textit{marginal} number of distinct ideas against prior egos (and not future egos), the ideas of the treatment egos remain non-redundant more successfully than the control egos once all egos complete the study. 

Ultimately, at a network level, the treatment egos collectively accumulate significantly more distinct ideas per round than their control counterparts ($\beta = 8.70, SE = 1.78, P<0.001$; linear mixed-effects model; Supplementary Table~\ref{groupwise_t2}).

Both marginal distinct idea count and non-redundant idea count metrics assess the novelty of an idea based on how `rare' it is against socially generated idea pools. This social rarity-based scoring approach falls short in accounting for an idea's semantic and subjective qualities. Therefore, to complement the above analysis, we next examine the semantic properties of the egos' ideas using the Creativity Quotient (CQ) metric. CQ employs information-theoretic measures to quantify creativity based on the semantic diversity of one's idea set (higher CQ indicates better creativity; see Methods). We find that the treatment egos had significantly higher CQ than the control egos (Figure~\ref{performance_AI}(C); $\beta = 0.70, SE = 0.23, P<0.01$; linear mixed-effects model; Supplementary Table~\ref{cq_t2}). Interestingly, the AI never seeks to optimize the semantic qualities of the egos' ideas. Yet, the boost in CQ in the treatment condition suggests that the intervention helps the egos leverage a broader set of semantic categories in their ideas compared to the control condition.

Lastly, we look at the novelty patterns of the egos' ideas using SemDis scores. SemDis is a computational method validated to approximate human raters' subjective novelty assessments of ideas~\cite{beaty2021automating}. A higher SemDis score indicates a higher novelty of an idea. We find that the person-level best ideas of the treatment egos are significantly better than those of the control egos (Figure~\ref{performance_AI}(D); $\beta = 0.012, SE = 0.005, P<0.05$; linear mixed-effects model; Supplementary Table~\ref{semdis_t2}). Once again, the AI does not directly optimize subjective assessments of novelty, and this effect emerges as a byproduct. This finding is of practical importance since, in many idea-generation contexts, we may only care about picking the best ideas for further consideration.

Taken together, the results show that the AI intervention not only boosts the number of distinct ideas of the treatment egos---as intended---but also elevates their performance in non-redundancy, semantic diversity, and novelty measures.

\subsection*{AI-driven networks are more decentralized than AI-agnostic networks, but not perfectly egalitarian}

\begin{figure}
    \centering
    \includegraphics[width=1\linewidth]{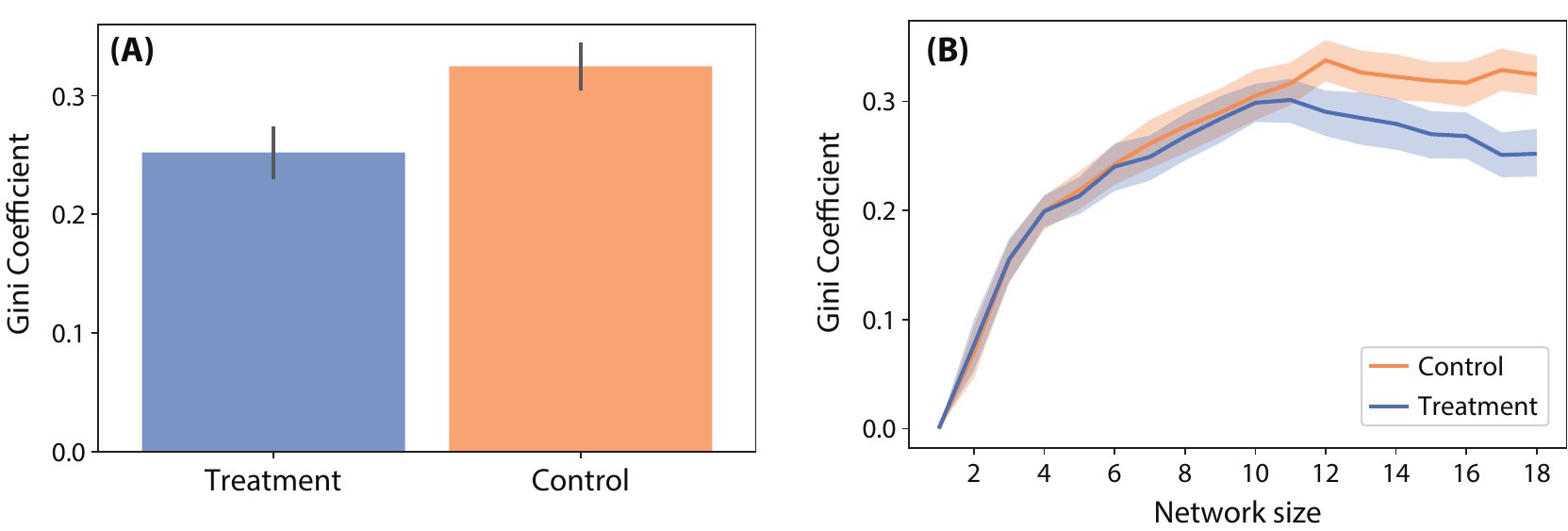}
    \caption{\textbf{Gini Coefficient analysis.} (A)~Gini Coefficients are significantly lower in the treatment condition than in the control. (B)~Gini Coefficients versus network size. The Gini Coefficients in the treatment condition drop significantly from the control condition at large network sizes. Whiskers and shaded regions denote $95\%$ C.I.}
    \label{gini_AI}
\end{figure}

The bipartite network among the egos and alters is initialized with a perfectly decentralized (egalitarian) structure, where all alters have the same number of ego-followers in the first round. Namely, as the egos join the experiment asynchronously, the network structure in Figure~\ref{generic_protocol}(A) is used to initialize their alter connections in round $1$. From there, the egos in both conditions choose their own alters to take inspiration from in the subsequent rounds. Although there are five rounds of idea generation, the egos submit their alter followee choices at the end of round $5$ as well, which gives us the network structures for a hypothetical sixth round. 

We employ the Gini Coefficient metric to probe how the inequality in the alters' follower counts evolves in rounds $2$ to $6$ as the egos make their network rewiring choices. In this metric, a perfectly decentralized network (i.e., one with uniform degree centrality across alters, as enforced in the first round) has a Gini coefficient of 0, and a higher Gini Coefficient denotes higher inequality in the alters' follower counts (see Methods). The treatment networks show significantly lower Gini Coefficients than the control networks (Figure~\ref{gini_AI}(A); $\beta = -0.07, SE = 0.01, P<0.001$; linear mixed-effects model; see Supplementary Table~\ref{gini}). In other words, the treatment egos `spread out' their inspiration sources more than the control egos. However, the treatment condition's Gini Coefficients are still significantly larger than zero, implying that the treatment networks are not perfectly egalitarian. Prior findings suggest that such \textit{partial} spreading out of the egos' inspiration sources can help reduce inter-ego idea redundancy (by reducing the overlap in people's inspiration sources) while retaining sufficient stimulation opportunities from the top ideators' ideas~\cite{baten2022novel}. This may partly explain the higher creative performance of the treatment egos we observe.

Upon closer inspection, we find that the Gini Coefficients of the treatment networks reduce significantly compared to the control networks only when the network sizes grow larger than ten egos in the respective rounds (Figure~\ref{gini_AI}(B)). This intuitively makes sense since, at larger network sizes, the pool of ideas collectively submitted by the egos becomes large enough to introduce inter-ego redundancy substantially. Lowering Gini Coefficients (and thereby spreading out inspiration sources) at larger network sizes can thus have a pronounced effect in reducing inter-ego redundancy and elevating creative performances.

\subsection*{Network structural features impact AI decision-making more in larger networks} 

\begin{figure}[t]
    \centering
    \includegraphics[width=1\linewidth]{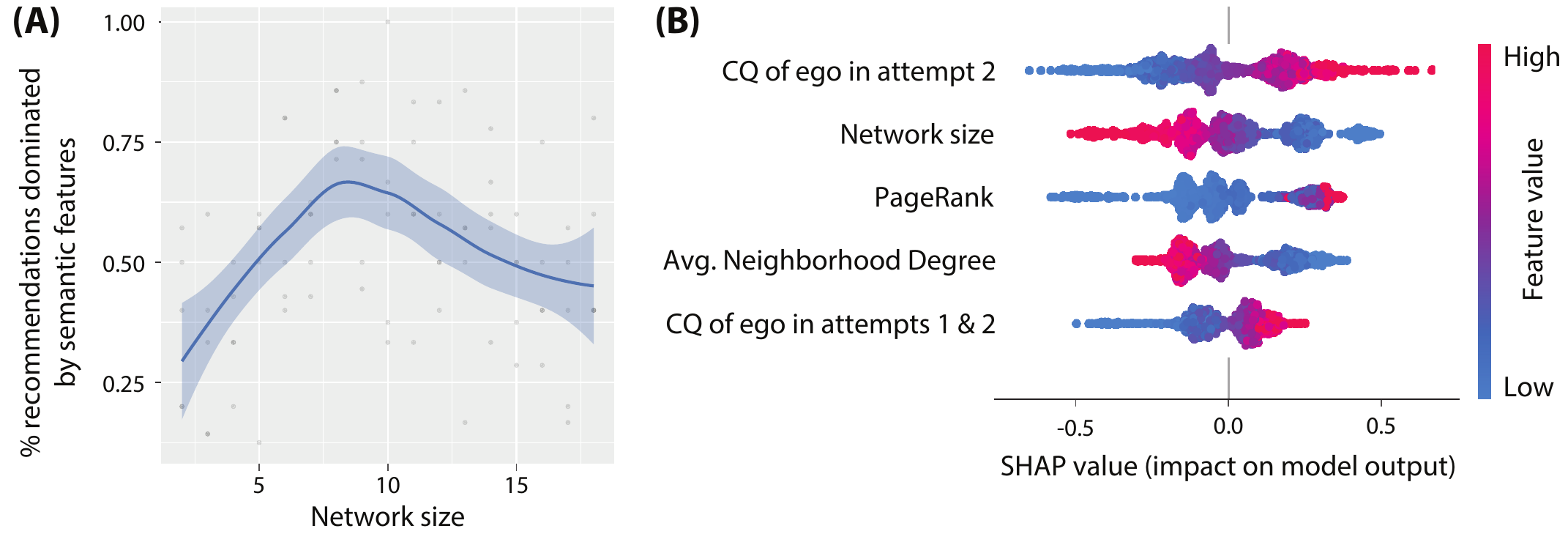}
    \caption{\textbf{Model diagnosis.} (A)~Fraction of AI recommendations dominated by semantic features across different network sizes. At each network size for each round, we calculate the fraction of decisions among all trials where the semantic features dominated the AI decision-making instead of network-structural features. Feature domination is determined by identifying which feature category played the most influential role in making the best alter-pair choice better than the alternatives. The solid line is estimated through Local Polynomial Regression Fitting. The shaded region denotes $95\%$ C.I. (B)~SHAP values show the individual contributions and relationship directions of the top $5$ features on the performance score prediction outputs.}
    \label{frac_sem}
\end{figure}

The performance score prediction model in \texttt{SocialMuse} employs two broad categories of input features: semantic and network structural features. We use SHapley Additive exPlanation (SHAP) values~\cite{lundberg2020local,scott2017unified} to probe how the two categories of features affect the AI-driven peer recommendations. SHAP values help attribute a prediction to different features and quantify the impact of each feature on the model output. Based on SHAP values, we identify the feature that played the most influential role in making the best alter-pair choice better than other alternatives. We then analyze these dominant features at a category level (see Methods).

Figure~\ref{frac_sem}(A) shows the fraction of AI-generated recommendations dominated by semantic features across different network sizes. We consider network sizes $2$ to $18$ for this analysis since comparing the impacts of the two categories makes less sense for a network size of $1$ (which has no meaningful network structural information). As the network size grows, the dominance of the semantic category depicts an inverse U-shaped trend, peaking at around the network size of $8$. Beyond this peak, the network structural features increasingly dominate AI decision-making in larger networks. In other words, the AI increasingly focuses on the network-structural features as many egos join the network (and the pool of ideas grows bigger). The growing importance of network-structural features in large networks may partly explain a mechanism behind lowering the Gini Coefficient at similarly large network sizes in the treatment condition. We also conduct ablation experiments to probe the relative importance of the two feature categories. Using only the semantic features in the model leads to a test set performance of $R^2=10.80\%$, whereas using only the network structural features leads to a test set performance of $R^2=26.31\%$. This illustrates the importance of the network structural features in the model performance.

The Beeswarm plot in Figure~\ref{frac_sem}(B) shows the individual contributions and relationship directions of the top $5$ features on the performance score prediction outputs. The Creativity Quotients of an ego's ideasets strongly and positively influence the model output, i.e., if an ego generates semantically diverse ideas in the current round, they are likely to score highly in the marginal distinct idea count metric in the next round. Network size impacts the model output negatively, intuitively suggesting that at larger network sizes, it becomes challenging to generate novel distinct ideas. If the ego enjoys a high PageRank centrality in the one-mode projected graph on the egos---i.e., a graph where two egos are connected by a weighted edge denoting the number of common alters they follow---the ego is more likely to have a high predicted score. In contrast, having a high average neighborhood degree in the projected graph harms the predicted performance. 

We provide the Beeswarm plots of the ablation scenarios in Supplementary Figure~\ref{frac_comb}. In addition to the insights above, it is evident that the semantic distances between the ego and alters' idea sets strongly influence the prediction outcomes. In the one-mode projected graph on the egos, lower global clustering coefficients correspond to higher predicted scores, i.e., it is difficult for ideas to stand out when the projected network becomes too clustered. Closeness centrality shows an inverted U-shaped relation with the performance score, implying that having too high or too low closeness centrality in the projected graph can diminish one's predicted performance scores. These insights provide further nuances on how the different categories of features affected the AI model decisions.

\subsection*{Being recommended boosts one's novelty ratings} 

In the treatment condition, different alters can be recommended to different egos for the same future round based on the differences in relevant contextual factors. Furthermore, the alters' ideas are shown to all egos identically in the rewiring stage of each round, where the egos rate all of the shown ideas on novelty before submitting their alter-followee choices. This setting allows us to systematically probe whether the recommendations have any causal impact on the egos' perception of the alters' ideas. To test this, we analyze whether the same idea from the same alter receives any different novelty ratings from the treatment egos when the alter is recommended versus when they are not recommended. We find that being recommended significantly boosts novelty ratings received by an alter's ideas ($\beta = 0.042, SE=0.013, P<0.001$; linear mixed-effects model; see Supplementary Table~\ref{rating}). In other words, the egos perceive an alter's ideas to be more \textit{novel} when they are primed by the \textit{peer recommendations} on the screen. The effect of priming has been widely studied in behavioral psychology~\cite{molden2014understanding,kahneman2011thinking}, but we are unaware of prior scientific evidence that peer recommendations can bias people's perception of idea novelty. Importantly, this finding paints a picture of how the AI-generated recommendations influence ego behavior in the experiment.

\section*{Discussion}\label{discussion}

Our findings contribute to (i)~the scientific study of creativity in self-organizing social networks and (ii)~the practical scope of building intelligent peer recommendation systems to elevate creativity in this setting. We trained a supervised learning model that predicts ideation performances based on semantic and network structural features. Using this model, we built a peer recommendation engine, \texttt{SocialMuse}, that maximizes people's predicted performance scores to make peer recommendations. Through a randomized controlled experiment, we showed that AI-informed treatment networks leveraging \texttt{SocialMuse} significantly outperformed AI-agnostic control networks on several measures of creativity. We further found that the treatment networks were more decentralized at large network sizes than control networks, partly due to the AI model increasingly prioritizing network-structural features at large network sizes.

A key challenge in the intervention design was to balance maximizing idea stimulation and eliminating idea redundancy. Prior findings suggest that dispersing the centrality of the highest performers to a ``Goldilocks'' extent (not too hot, not too cold, just right) can help strike such a balance---allowing people to benefit from taking inspiration from the most creative people's ideas but also slightly spreading out who people take their inspirations from~\cite{baten2022novel,baten2020creativity}. Contrarily, networks whose dispersal is too narrow or too broad are shown to lead to inferior outcomes. However, no analytic approach is yet known to solve for an optimal global network structure that balances these forces. In our work, we took a local (myopic) optimization approach where the AI, at a time, only attempted to maximize a single participant's idea generation performances in a single future round based on the contextual factors surrounding the participant. We showed that such local optimization of network connections can contribute to partially decentralizing the network and lead to ideation advantages. However, given the lack of analytic backing, we cannot comment on how well this approach approximated the theoretical limits of possible creativity benefits.

A second challenge in the intervention design was the constant change in the information environments as people self-organized in the network. Namely, people's idea generation characteristics and peer choices may take many paths over time as peer interactions occur in a complex social system. Consequently, the underlying theoretically optimal network structures may change as the broader social system evolves. Our local optimization approach navigated around this challenge by allowing us the adaptability to use a single trained model for all participants in all rounds as the social networks evolved temporally. 

Taken together, our results (i)~corroborate prior findings that partial decentralization of social networks can elevate the creative performances of its members and (ii)~extend prior understanding of the scope by showing that a local, data-driven optimization approach can offer a practical way of decentralizing the networks, particularly in the context of evolving information environments. These insights can help advance the theory of collective creativity in temporal social networks~\cite{acar2023collective,perry2003social,jiang2018dynamic}.

In terms of practical application building, our study contributes to the growing body of intelligent recommendation systems research. Peer recommendation (e.g., on Facebook or Twitter) and content recommendation (e.g., on Pinterest or YouTube) engines have become cornerstones of the modern internet experience. Peer recommendation systems commonly rely on homophily, where people are suggested connections based on the number of common neighbors~\cite{hossain2021fast} or shared features such as age, race, location, job, lifestyle, influence, or trust~\cite{Ning2019personet,cheng2019friend,proma2024exploring}. Closer to our scope, people increasingly seek idea inspiration from peers on online platforms like Behance, Pinterest, or ResearchGate. However, existing peer recommendation systems typically optimize user engagement and user satisfaction metrics rather than the users' creative outcomes~\cite{he2016vista,ziegler2005improving,kamath2013board,kislyuk2015human}. Our lab experiment-based results provide proof of concept that such platforms can help their users be more creative by recommending peers intelligently. We list below a few implications on how online platforms can leverage our findings:

\textit{First,} the operationalization of semantic and network context features will differ based on the real-life platform's specific offerings and characteristics. We recorded ideas in text format in our experiment, allowing us to use natural language processing and information-theoretic methods to analyze the semantic properties of the ideas. This approach can be leveraged in platforms with text-heavy content, e.g., academic research-centric social media (e.g., ResearchGate), and teamwork-centric online platforms (e.g., Slack). In contrast, platforms for photographers and audio-visual artists, for instance, may instead need to leverage computer vision and audio signal processing-based methods to compute semantic features. Similarly, we constructed network features from a bipartite network, whereas real-life platforms may need to calculate the features from one-mode social networks. The exact feature importance insights (i.e., SHAP values and directions of features) may thus vary across platforms and application settings. What constitutes `creative' performance will also vary across application settings. Our work provides general nuances and insights for balancing the semantic and network structural context factors for ideation benefits in those settings.

\textit{Second,} we found an XGBoost-based machine learning model to perform the best in learning complex, non-linear interactions among the multifaceted input features directly from data. XGBoost employs an ensemble-based boosting approach that boasts efficient regularization, better parallel processing, cache access, data compression, and better handling of sparse data, among other benefits~\cite{chen2016xgboost,han2012data}. In particular, our small-scale data setting warranted such machine learning models. Large neural models may be more suitable in industry-scale data settings for obtaining superior prediction performances~\cite{baten2023predicting}.

\textit{Third,} given the small search space of alter-pair alternatives, we could use a brute force search approach to consider all possible options before generating the recommendations and still finish all computations in near real-time during deployment. This was sufficient for answering our research questions. However, despite its comprehensive search capabilities that lower prediction errors, brute-force search algorithms will not scale well for massive networks with millions of users. In such settings, a two-step process is usually necessary: (i)~narrowing the millions of users to a smaller set of candidate users for further consideration (\textit{retrieval} step) and then (ii)~ranking the candidate subset based on appropriate heuristics to identify the final peer recommendation list (\textit{ranking} step)~\cite{covington2016deep,kolodner2024robust}. Our results offer heuristics that may help both of these steps toward developing industry-scale peer search solutions.

\textit{Fourth,} in our bipartite network structure, the egos took inspiration from the alters and never interacted with other egos. However, some of our metrics evaluated the egos' performance against other egos (and not against alters). In real-life settings, people may not know all of the prior art in their respective domains, which is analogous to the ideas of the `hidden' egos our participants competed against for novelty. Indeed, people may browse online platforms for inspiration based on their needs, making the inspiration-seeking activity inherently asynchronous, as was true for the egos in our experiment. People may not know every prior person who generated ideas on the task and who those prior ideators took inspiration from---mimicking our focal egos' ignorance of the network choices of prior egos. Our bipartite setting allowed us to systematically separate inspiration sources and stimulation outcomes to better understand the mechanisms underlying creativity in complex systems. Our insights imply that in real-life, one-mode networks, designing interventions to partially disperse the high centrality/visibility of top ideators (e.g., academics with high $h$-indices or artists with massive following in online platforms~\cite{azoulay2010superstar,salganik2006experimental}) may benefit the creative qualities of the inspired ideas in the social systems.

\textit{Fifth,} we installed several scaffolding mechanisms throughout the experiment, such as asking egos to justify their alter choices every round and having a recall test at the end of the experiment. These were in place to (i)~ensure that the participants paid attention to the stimuli~\cite{nijstad2006group,dugosh2005cognitive,paulus2000groups,brown1998modeling}, (ii)~raise their epistemic motivation, and (iii)~improve their systematic information processing~\cite{bechtoldt2010motivated,scholten2007motivated}. Furthermore, we associated the recommendations with SHAP-value-based explanations to inform the egos of the recommendation rationales. In real-world application settings, having similar measures can help accentuate the stimulation benefits of AI-driven peer recommendations.

Recent literature on the intersection of AI and creativity paints a `cautiously optimistic' picture. On the one hand, large language models (LLMs) and other generative AI technologies promise to boost people's productivity in creative work~\cite{wingstrom2024redefining}. On the other hand, an over-reliance on generative AI can hurt creativity and critical thinking, particularly in higher education~\cite{ivanov2023dark}. A recent analysis shows that humans increasingly imitate LLMs in their spoken language~\cite{yakura2024empirical}---which suggests that being implicitly primed by LLMs may potentially limit the diversity of words and concepts people access during creative ideation. Our findings add a positive use case of AI to this discourse, showing how AI can proactively boost a social system's creative outcomes by taking the role of a peer recommendation engine rather than being a generative source of inspiration.

At the same time, our work brings to the fore questions of transparency, trustworthiness, fairness, and accountability in AI systems. Despite achieving enormous success in solving practical problems, data-driven intelligent systems are often criticized for abusing private information, delivering biased outputs, and perpetuating harm at individual and social levels~\cite{dignum2019responsible}. In the context of our objectives, poorly designed implementations of peer recommendation systems can perpetuate bias and harm similarly. A lack of transparency in generating recommendations can also decrease people's trust in such systems. We found evidence that psychologically priming people with \textit{recommendations} can influence their perception of the \textit{novelty} of ideas---further underscoring the point that people may not always realize how these AI systems can affect their behavior. Ensuring explainability and proper consent mechanisms can help alleviate some of the concerns in this scope. As the field advances, it is crucial to analyze the selection of input features critically, consider how these may or may not connect to the user's rights, and frequently re-evaluate the design of such AI systems.

\section*{Methods}\label{methods}

\subsection*{\texttt{SocialMuse} System}
\textbf{Performance score prediction model training.} \textit{Training data.} We trained a regression-based prediction model using the data from three of our previous experiments~\cite{baten2020creativity,baten2021cues,baten2022novel}. Namely, we curated data from the (1)~`Dynamic' condition of~\cite{baten2020creativity}, (2)~both conditions of~\cite{baten2021cues}, and (3)~`Baseline’ condition of~\cite{baten2022novel}. Each prior experiment had several independent networks of 18 egos, leading to a total of $N_{\text{prior}}=360$ egos in the curated dataset. The egos asynchronously completed their tasks of generating ideas on the five prompt objects in five rounds (the prompt objects and their sequence were consistent across experiments). The ideas in the dataset are annotated so that the same ideas are assigned the same unique idea ID.

\textit{Target score.} We reanalyzed the prior data to calculate the `marginal distinct idea count' of each ego in each round as the target score in our prediction. To capture this, we added the egos to their $18$-ego networks one at a time (in the order in which they completed the original experiments). This generated a growing pool of ideas for each round as the egos were added to the network. Each ego was scored on the number of new ideas they contributed to this pool in attempt 2 of each round. Thus, each ego $e_i$ contributed four data points with target scores $s_{e_i}^t$ capturing their marginal distinct idea counts in rounds $t \in \{2, 3, 4, 5\}$.

\textit{Input features.} As input features of these data points, we considered (i)~Semantic and (ii)~Network structural features. 

\begin{enumerate}[(i)]
    \item Semantic features:
    \begin{itemize}
        \item Ego's performance features: From round $t-1$, we calculated ego $e_i$'s Creativity Quotients (defined later) from their attempt-1 idea set, attempt-2 idea set, and the combined idea set from both attempts, giving us three features.
        \item Alters' performance feature: We considered the two alters that the ego followed in round $t$. From round $t-1$, we took the summation of Creativity Quotients of those two alters as an input feature.
        \item Ego-alter and alter-alter semantic distances: From round $t-1$, we considered the idea sets of the two aforementioned alters and the attempt-1 idea set of the ego. We computed the semantic distances between the idea sets of the ego and each alter, from which we took the max, min, mean, and standard deviation of the semantic distances. We also combined the idea sets of the two alters to create one document and calculated the semantic distance between the ego's idea set and the larger concatenated document. This gave us four semantic distance measures (min, max, mean, standard deviation, and overall distance). Additionally, we considered the semantic distance between the two alters' idea sets. We used three approaches for calculating the semantic distances: (i)~Cosine similarity using Word2Vec embeddings~\citep{mikolov2013efficient}, (ii)~Word Movers' Distance using Word2Vec embeddings~\citep{kusner2015word}, and (iii)~Cosine similarity using GloVe embeddings~\citep{pennington2014glove}. This gave us a total of $18$ input features.
        \item Gender diversity: We computed how many of the aforementioned alters were of a different gender than the ego. This is based on earlier findings that the semantic properties of people's ideas can be associated with their demographic identities, such as gender~\cite{baten2021cues}.

    \end{itemize}
    \item Network structural features: We considered the bipartite network upto and including the focal ego in round $t$, from which we collected the following features:
    \begin{itemize}
        \item Network size
        \item Gini Coefficient of all of the alters' follower counts
        \item Global clustering coefficient and transitivity of the weighted undirected one-mode graph projected on the egos
        \item Local clustering coefficient of the ego in the weighted undirected one-mode graph projected on the egos
        \item Degree, betweenness, Eigenvector, closeness, and PageRank centralities of the ego in the weighted undirected one-mode graph projected on the egos
        \item Average neighborhood degree of the ego in the weighted undirected one-mode graph projected on the egos
        \item The number of triangles that include the ego as a constituting node in the weighted undirected one-mode graph projected on the egos.
    \end{itemize}
\end{enumerate}
Lastly, we included a round ID feature to capture trends specific to the prompt of round $t$.

\textit{Model training.} We split the data into training and test sets using a $80:20$ ratio. To prevent information leakage, we ensured that data points from the same ego remained in the same split. We used the Boosted Recursive Feature Elimination (BoostRFE) method implemented in the \texttt{shap-hypertune} Python package to select features based on their SHAP importance. We standardized the features using the StandardScaler method implemented in the \texttt{sklearn} Python package. We trained the following regression models: XGBoost, LightGBM, Support Vector Regression (SVR), RandomForest, and Ridge. We used grid search with $5$-fold cross-validation to tune hyperparameters (see Supplementary Table~\ref{hyperparameters}). We found an XGBoost model ($\mathcal{M}$) to show the best test set performance of $R^2 = 32.58\%$ (see Supplementary Table~\ref{model_results}).

\textbf{Recommendation generation module.} \textit{Identifying the best alter-pair.} During deployment, at the end of round $t-1$, \texttt{SocialMuse} recommended an ego two alters to follow (out of six) in the upcoming round $t$. Each trial in our validation experiment had $n_a=6$ alters, and the ego needed to follow exactly $k=2$ of them at the end of each round. This lead to $\binom{n_a}{k}=15$ alter-pair alternatives. Given the small search space, we took a brute-force search approach to find and recommend the most promising alter-pair to each ego in each round. To operationalize this, we constructed $15$ sets of network features from $15$ `hypothetical' network structure alternatives that would emerge if the ego followed the different alter-pair options. Namely, for each alter-pair alternative, we added the two ego-alter connections to the network structure emerging from the connection choices of prior egos in round $t$. We computed the network-structural features (the ones listed above) from that hypothetical network. We also computed semantic features from round $t-1$ for each of those $15$ configurations (same features as listed above). Each set of input features was then run through $\mathcal{M}$ to predict the ego's marginal distinct idea count scores in round $t$. Finally, we recommended the highest-scoring alter-pair to the ego. 

\textit{Generating SHAP-based explanations for the recommendations.} We used SHapley Additive exPlanation (SHAP) values~\cite{lundberg2020local,scott2017unified} to identify which feature exerted the most influence in making the best alter-pair a better choice than other alternatives and accordingly showed the ego an explanation behind the AI-generated recommendation. SHAP values are helpful for intuitively explaining the outputs of supervised machine learning models. Building on cooperative game theory, SHAP computes a contribution score for each feature, denoting the feature's impact on the model's output. These contribution scores are calculated by assessing the feature's value in relation to all possible combinations of features in the input data. We tracked the absolute SHAP values of all features for each of the $15$ alter-pair predictions made using $\mathcal{M}$. For each feature, we calculated the absolute difference between the best-alter-pair predictions's absolute SHAP value and the other $14$ alter-pair predictions' mean absolute SHAP value. We identified the feature having the largest absolute difference in this setting. If the feature belonged to the semantic features category, we showed the explanation ``Recommended for better inspiration" to the ego; otherwise, we showed ``Recommended for reducing idea redundancy" to the ego.

\subsection*{Validation Experiment}
\subsubsection*{Participants}
 We recruited $420$ US-based participants from Prolific. There were $10$ trials with independent participants. Each trial had $6$ alters, $18$ egos in the treatment condition, and $18$ egos in the control condition. Participants were randomly assigned their roles (alter/ego) and study conditions. The race distribution was: White: $314$, Black or African American: $53$, Asian: $22$, two or more races: $7$, other: $24$. The gender distribution was: female: $177$, male: $238$, other: $5$. The age distribution was: 18-24: $45$, 25-34: $178$, 35-44: $116$, 45-54: $51$, 55+: $30$. Two-sample Chi-squared tests did not show any detectable difference in the race, gender, or age distributions across the control and treatment conditions ($P>0.05$ for each demographic variable). Thus, any observed differences among the study conditions cannot be systematically attributed to differences in the distributions of the demographic categories.

\subsubsection*{Procedure}
\textbf{Creativity task.} Unlike convergent thinking, which requires individuals to zero in on known correct answers (as tested in traditional school exams), divergent thinking leads people to generate \textit{numerous} and \textit{varied} responses to a given prompt or situation~\cite{runco2014creativity,kozbelt2010theories}. Our divergent thinking task is based on the canonical Alternate Uses Test\footnote{(Guilford's Alternate Uses Test is Copyright @ 1960 by Sheridan Supply Co., all rights reserved in all media, and is published by Mind Garden, Inc, www.mindgarden.com)}~\cite{guildford1978alternate}. In each of the five rounds, the participants were instructed to consider an everyday object (e.g., a brick) whose common use was stated (e.g., a brick is used for building). We chose the first five objects from Form B of Guilford's test as the prompt objects for the five rounds. The participants needed to generate alternative use ideas that were novel and appropriate, different from each other, and different from the given common use. The participants were guided by examples specified in the test manual.

\textbf{Protocol.} In each trial, we first recorded the ideas of the $6$ alters (single attempt in 3 minutes). Then, we used those alters' ideas as stimuli for the egos of both conditions in that trial. The egos in each condition were randomly assigned network positions using the initial structure shown in Figure~\ref{generic_protocol}(A), where each ego was connected to $2$ alters out of the $6$ in the trial. Each round had three stages: First, the egos generated ideas independently (attempt 1). Second, the egos were shown the ideas of the $2$ alters they were following. They could submit any new ideas that were inspired by the alters' ideas (attempt 2). Third, the egos were shown the ideas of all $6$ alters, which they rated on novelty on a 5-point Likert scale (1: not novel, 5: highly novel). Then, the egos could optionally choose for themselves which $2$ alters (out of $6$ alters) to follow in the next round (Figure~\ref{generic_protocol}(B)). Since every ego rated the ideas of \textit{all 6 alters} before making the following/unfollowing choices, they all had the same global knowledge about the alters' ideas. Thus, we do not anticipate bias in the egos' following patterns from varied exposure to the alters.

The egos were given $3$ minutes for each of their attempts in each round. They could not directly re-submit their followee alters' ideas as their own and were informed that their number of non-redundant ideas would contribute to their performance. The egos were informed that they would have a short test at the end of the study, asking them to recall ideas shown to them. The purpose of the recall test was to ensure the participants' attention to the stimuli ideas, which is known to have positive stimulation benefits~\cite{nijstad2006group,dugosh2005cognitive,paulus2000groups,brown1998modeling}. As the egos rewired connections in each round, they were required to submit the rationale behind their choices. This helped ensure the accountability of the egos in making their choices, which is known to raise epistemic motivation and improve systematic information processing~\cite{bechtoldt2010motivated,scholten2007motivated}. The participants were paid \$10 upon completing the tasks and a bonus of \$5 if they were among the top 5 performers in their trials in terms of non-redundant idea counts.

\subsubsection*{Measures}

\textbf{Measuring creative performance.} We used four complementary metrics for quantifying creative performances, as commonly employed in the literature: marginal distinct idea count, non-redundant idea count, Creativity Quotient, and SemDis scores. The first two metrics quantify how rare one's ideas are compared to the peers' ideas~\cite{oppezzo2014give,abdullah2016shining}, but do not assess the ideas' intrinsic qualities (i.e., even a great idea is not novel in these metrics if many people submit it). In contrast, the Creativity Quotient metric captures how semantically diverse a person's idea-set is~\cite{snyder2004creativity,bossomaier2009semantic} but does not attempt to compare idea-sets socially (i.e., two people with highly diverse yet identical idea-sets will achieve identically high scores). Finally, SemDis scores computationally approximate the subjective notion in creativity assessment.

\textit{Marginal distinct idea counts and non-redundant idea counts.} We first discarded inappropriate submissions that did not meet the specified requirements. Since the same idea can be phrased differently by different people, we collected or `binned' the same yet differently phrased ideas together under common bin IDs. For binning the ideas, we followed the coding rules described by Bouchard and Hare~\cite{bouchard1970size} and the scoring key of Guilford's test. We scored the egos on marginal distinct idea counts and non-redundant idea counts using these bin IDs as described previously. 
For collective-level analysis, we took the total number of distinct bin IDs generated by the egos in a given condition in a given trial as the collective performance marker.

The first author binned all the ideas in the dataset, while two other research assistants independently binned the ideas of a random $25\%$ of the participants. The annotators were shown the ideas in a random order. Based on their independently curated bin IDs, we computed each participant's total non-redundant idea counts in all $5$ rounds together and calculated the agreements among the coders. The agreements were high both between the first and second coder (intra-class correlation coefficient, $ICC(2,2)=0.92$, $P<10^{-32}$, $95\%$ C.I. = $[0.89,0.95]$; Pearson's $r=0.86$, $P<10^{-31}$, $95\%$ C.I.=$[0.80,0.90]$), and between the first and third coder (intra-class correlation coefficient $ICC(2,2)=0.89$, $P<10^{-25}$, $95\%$ C.I. = $[0.84,0.93]$; Pearson's $r=0.82$, $P<10^{-26}$, $95\%$ C.I.=$[0.75,0.88]$). We used the bin ID annotations from the first coder in the analyses.

\textit{Creativity Quotient (CQ).} This metric builds on the intuition that if a participant's ideas are very similar to each other, they are likely subtle variations of a small number of semantic categories. Conversely, if the ideas are dissimilar, they likely touched many semantic categories---marking better creativity~\cite{snyder2004creativity,bossomaier2009semantic,rietzschel2007personal}. CQ employs an information-theoretic semantic similarity measure derived from WordNet~\cite{miller1995wordnet} to capture this intuition. 

Concepts are organized as syn-sets or synonym sets in WordNet, where the nouns are linked with `is a' relationships. We removed stop-words and punctuation from the ideas and ran a spell-checker. Next, we split the ideas into their constituting set of concepts and converted the terms into nouns for availing the `is a' relationships. We then computed the information content of those concepts. The taxonomic organization of WordNet implies that concepts with many hyponyms convey less information than concepts with fewer hyponyms~\cite{seco2004intrinsic}. Therefore, infrequent concepts at the leaf nodes hold more information than their abstracting nodes. The Information Content, $I$, of a concept $c$ can thus be calculated as,
\begin{equation}
    I(c) = \frac{log\big(\frac{h(c)+1}{w}\big)}{log\big(\frac{1}{w}\big)} = 1- \frac{log(h(c)+1)}{log(w)},
\end{equation}
where $h(c)$ is the number of hyponyms of $c$, and $w$ is the total number of concepts in WordNet. The denominator normalizes the metric against the most informative concept to ensure $I\in[0,1]$.

Next, we calculated the semantic similarity of each pair of concepts in the idea-set, $c_1$ and $c_2$, as~\cite{jiang1997semantic},
\begin{equation}
    sim(c_1,c_2)= 1-\Big(\frac{I(c_1)+I(c_2)-2\times sim_{MSCA}(c_1,c_2)}{2} \Big),
\end{equation}
where $sim(c_1,c_2)$ is a function of the information overlap between the two concepts, $sim_{MSCA}(c_1,c_2)$. This overlap is calculated using the information content of the Most Specific Common Abstraction (MSCA) that subsumes both concepts,
\begin{equation}
    sim_{MSCA}(c_1,c_2) = \max_{c'\in S(c_1,c_2)} I(c'),
\end{equation}
where $S(c_1,c_2)$ is the set of concepts subsuming $c_1$ and $c_2$.

Given the pair-wise concept similarities, we computed the multi-information, $I_m$, as the shared information across the idea set. We crafted the max spanning tree from the network of concepts and their pairwise similarity values. We summed over the edge weights in the max spanning tree to get $I_m$. Finally, we calculated $Q$ as,
\begin{equation}
    Q = N-I_m,
\end{equation}
where $N$ is the total number of concepts in the person's idea set. 

\textit{SemDis scores.} The open platform SemDis~\cite{beaty2021automating} approximates human subjective judgments of an idea's novelty using natural language processing-based computational methods. We computed the SemDis scores of each idea in our dataset using the publicly available online SemDis tool\footnote{http://semdis.wlu.psu.edu/}.

\textbf{Measuring inequality in the alters' follower counts.} The popularity of an alter $i$ is defined by his/her share of followers, $m_i = d_i/ \sum_{k=1}^{S}d_k$, where $d_i$ is alter $i$’s follower count in a given round and $S$ is the number of alters in the trial. The Gini Coefficient is then calculated by, 

\begin{equation}
    G = \frac{\sum_{i=1}^S \sum_{j=1}^S |m_i - m_j| }{2S \sum_{k=1}^S m_k}.
\end{equation}

This represents the average difference in follower counts between pairs of alters, normalized to fall between $0$ (complete decentralization) and $1$ (maximum centralization).

\backmatter
\bmhead{Acknowledgements}
Grants XX supported the study.

\bmhead{Conflicts of Interest} The authors declare no conflict of interest.
\bmhead{Ethics Approval} The study was reviewed and approved by the Institutional Review Board of the University of Rochester, NY, USA. All participants provided informed consent.
\bmhead{Data and Code Availability} The data and code are available on GitHub at \url{https://github.com/ROC-HCI/AI-creativity-social-network}. To comply with the copyright restrictions of the tasks, we publish the processed data from the creativity tasks rather than raw textual data.

\bmhead{Author Contributions} RAB designed and performed the research and wrote the manuscript. ASB and KV performed research. GG and EH designed and oversaw the research and wrote the manuscript.


\clearpage
\begin{appendices}
\renewcommand{\thefigure}{S\arabic{figure}}
\renewcommand{\thetable}{S\arabic{table}}

\section*{Supplementary Materials for \textit{AI can Enhance Creativity in Social Networks}}\label{SM}
\subsection*{Supplementary Text}

\textbf{Cognitive basis of creativity.} The classic Associative Theory posits that people's semantic memory (a long-term memory system) stores concepts in a meaningful way, where related concepts are more strongly connected than unrelated ones~\cite{santanen2000cognitive,patterson2007you,kenett2023creatively,kumar2021semantic,jones2015models}. Given an idea-generation prompt, a person's cognitive processes scan the local neighborhood of relevant concepts in their semantic memory and help generate an initial set of ideas~\cite{mednick1962associative}. After depleting these initial obvious ideas, a person can arrive at original ideas by accessing \textit{remote} concepts and recombining various aspects of those non-obvious concepts into novelty~\cite{runco2014creativity,nijstad2006group,volle2018associative}. This long-term memory circuitry can be stimulated by external priming, triggering additional ideas one could not think of on their own~\cite{wang2010idea,brown2002making,siangliulue2015toward}. This theoretical grounding suggests that a person's chances of encountering non-obvious stimuli can increase if they are recommended to `follow' peers whose ideas are semantically distant from their own ideas.

\clearpage

\subsection*{Supplementary Figures}

\begin{figure}[h]
    \centering
    \includegraphics[width=1\linewidth]{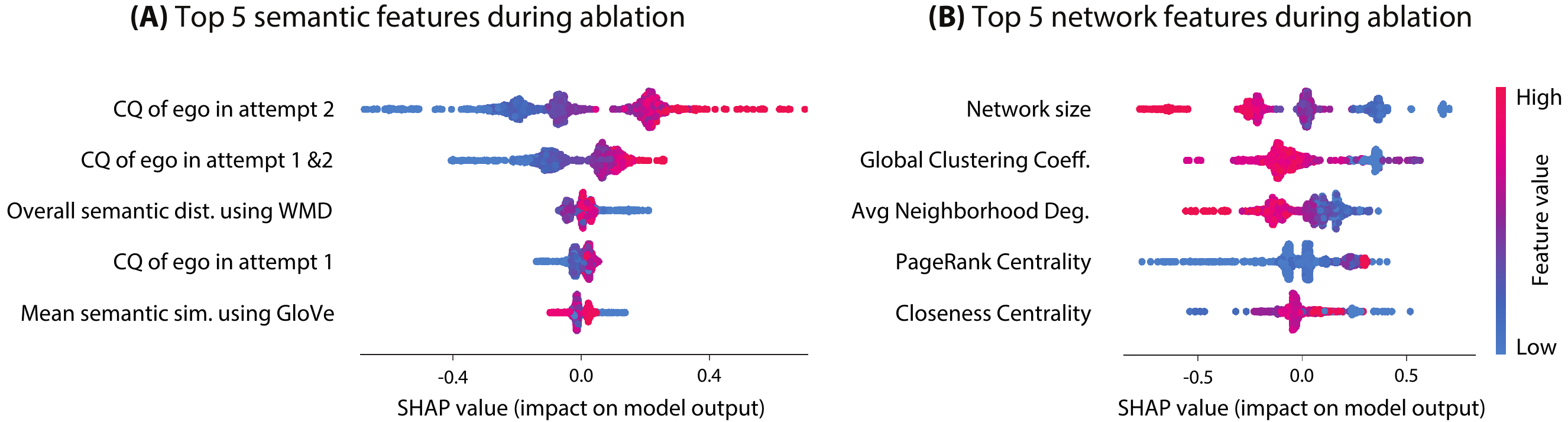}
    \caption{\textbf{Ablation insights.} The Beeswarm plots of SHAP values denote individual contributions and relationship directions of features in predicting marginal distinct idea counts. (A)~Top 5 features when the model is trained only on semantic features. A high Creativity Quotient of the ego corresponds to a high predicted score. In addition, the semantic distances and similarities between the ego and alters' idea sets strongly influence the prediction outcomes. (B)~Top 5 features when the model is trained only on network structural features. Smaller network sizes correspond to higher predicted scores. In the one-mode projected graph on the egos, lower global clustering coefficients and lower average neighborhood degrees correspond to higher predicted scores. Closeness centrality shows an inverted U-shaped relation with the performance score---where having too high or too low closeness centrality in the projected graph can diminish one's predicted performance scores.}
    \label{frac_comb}
\end{figure}

\clearpage
\subsection*{Supplementary Tables}

\begin{table}[h]
\centering
\caption{Comparison of marginal non-redundant idea counts (per ego per round) between control and treatment conditions using a linear mixed-effects model fit by REML. The round ID, network size, and ego ID are captured as random effects. The results show that the treatment egos significantly outperformed the control egos after accounting for all repeated measures. $t$-tests use Satterthwaite's method. ***$P<0.001$.}
\begin{tabular}{lrrrrrr}
   & Estimate ($\beta$) & Std. Error & df & $t$ value  & Pr($>t$) & \\ 
\midrule
 (Intercept)            & $2.49$ & $0.21$ & $21.12$  & $11.70$  & $1.09e-10$ & *** \\    
 C(Condition)treatment  & $0.46$ & $0.10$ & $341.00$  & $4.54$  & $7.76e-6$ & *** \\   
\bottomrule
\end{tabular}
\label{marginal_nonredun_t2_art}
\end{table}

\begin{table}[h]
\centering
\caption{Linear mixed-effects model fit by REML. The response variable is the marginal non-redundant idea counts per ego per round. The fixed-effect predictors are condition (categorical), round number (numeric), and network size (numeric). The ego id (categorical) is modeled as a random effect variable. The results once again show that the treatment egos significantly outperformed control egos. In addition, we see that the marginal non-redundant idea counts dropped significantly with increasing network size but did not show any significant roundwise trend. $t$-tests use Satterthwaite's method. ***$P<0.001$.}
\begin{tabular}{lrrrrrr}
   & Estimate ($\beta$) & Std. Error & df & $t$ value  & Pr($>t$) & \\ 
\midrule
 (Intercept)            & $3.62$ & $0.15$ & $786.33$  & $24.03$  & $2e-16$ & *** \\    
 C(Condition)treatment              & $0.46$ & $0.10$ & $357.00$  & $4.38$  & $1.6e-5$ & *** \\   
  Round                 & $0.04$ & $0.03$ & $1079.00$ & $1.65$   & $ 0.09942$ &  \\    
   Network Size         & $-0.13$ & $0.01$ & $357.00$  & $-13.35$  & $2e-16$ & *** \\    
\bottomrule
\end{tabular}
\label{marginal_nonredun_t2_lmer}
\end{table}

\begin{table}[h]
\centering
\caption{Comparison of non-redundant idea counts (per ego per round) between control and treatment conditions using a linear mixed-effects model fit by REML. The round ID and ego ID are captured as random effects. The results show that the treatment egos significantly outperformed the control egos after accounting for all repeated measures. $t$-tests use Satterthwaite's method. ***$P<0.001$.}
\begin{tabular}{lrrrrrr}
   & Estimate ($\beta$) & Std. Error & df & $t$ value  & Pr($>t$) & \\ 
\midrule
 (Intercept)            & $1.70$ & $0.12$ & $6.11$  & $14.20$  & $6.53e-6$ & *** \\    
 C(Condition)treatment  & $0.40$ & $0.10$ & $358.00$  & $3.92$  & $0.000105$ & *** \\   
\bottomrule
\end{tabular}
\label{nonredun_t2}
\end{table}

\clearpage

\begin{table}[h]
\centering
\caption{Comparison of network-level total distinct idea counts (per round) between the control and treatment conditions using a linear mixed-effects model fit by REML. The round ID and trial ID are captured as random effects. The results show that the treatment networks significantly outperformed the control networks after accounting for all repeated measures. $t$-tests use Satterthwaite's method. ***$P<0.001$.}
\begin{tabular}{lrrrrrr}
   & Estimate ($\beta$) & Std. Error & df & $t$ value  & Pr($>t$) & \\ 
\midrule
 (Intercept)            & $55.23$ & $2.17$ & $9.43$  & $25.45$  & $5.16e-10$ & *** \\    
 C(Condition)treatment  & $8.70$ & $1.78$ & $66$  & $4.90$  & $6.47e-6$ & *** \\   
\bottomrule
\end{tabular}
\label{groupwise_t2}
\end{table}

\begin{table}[h]
\centering
\caption{Comparison of Creativity Quotients (per ego per round) between the control and treatment conditions using a linear mixed-effects model fit by REML. The round ID and ego ID are captured as random effects. The results show that the treatment egos significantly outperformed the control egos after accounting for all repeated measures. $t$-tests use Satterthwaite's method. **$P<0.01$.}
\begin{tabular}{lrrrrrr}
   & Estimate ($\beta$) & Std. Error & df & $t$ value  & Pr($>t$) & \\ 
\midrule
 (Intercept)            & $6.52$ & $0.23$ & $11.72$  & $28.79$  & $3.08e-12$ & *** \\    
 C(Condition)treatment  & $0.70$ & $0.23$ & $358.00$  & $2.98$  & $0.00308$ & ** \\   
\bottomrule
\end{tabular}
\label{cq_t2}
\end{table}

\begin{table}[h]
\centering
\caption{Comparison of the maximum SemDis scores (per ego per round) between the control and treatment conditions using a linear mixed-effects model fit by REML. In other words, for each ego in each round, the idea with the highest SemDis score is considered a data point. The round ID and ego ID are captured as random effects. The results show that the treatment egos significantly outperformed the control egos after accounting for all repeated measures. $t$-tests use Satterthwaite's method. *$P<0.05$.}
\begin{tabular}{lrrrrrr}
   & Estimate ($\beta$) & Std. Error & df & $t$ value  & Pr($>t$) & \\ 
\midrule
 (Intercept)            & $1.05$ & $0.009$ & $3.88$  & $120.82$  & $4.45e-8$ & *** \\    
 C(Condition)treatment  & $0.012$ & $0.005$ & $358.00$  & $2.35$  & $0.0193$ & * \\   
\bottomrule
\end{tabular}
\label{semdis_t2}
\end{table}

\clearpage

\begin{table}[h]
\centering
\caption{Comparison of Gini Coefficients (per round) between the control and treatment conditions using a linear mixed-effects model fit by REML. The round ID and trial ID are captured as random effects. The results show that the treatment networks had significantly lower Gini coefficients than control networks after accounting for all repeated measures. $t$-tests use Satterthwaite's method. ***$P<0.001$.}
\begin{tabular}{lrrrrrr}
   & Estimate ($\beta$) & Std. Error & df & $t$ value  & Pr($>t$) & \\ 
\midrule
 (Intercept)            & $0.32$ & $0.02$ & $11.71$  & $20.07$  & $1.98e-10$ & *** \\    
 C(Condition)treatment  & $-0.07$ & $0.01$ & $89.00$  & $-6.37$  & $7.95e-9$ & *** \\   
\bottomrule
\end{tabular}
\label{gini}
\end{table}

\begin{table}[h]
\centering
\caption{Comparison of idea-level ratings between two cases: when the idea-generating alter was recommended to the ego versus when the alter was not recommended. Linear mixed-effects model fit by REML. The idea ID, ego ID (rater), and alter ID (idea generator) are captured as random effects. The results show that being recommended significantly boosts idea ratings. $t$-tests use Satterthwaite's method. ***$P<0.001$.}
\begin{tabular}{lrrrrrr}
   & Estimate ($\beta$) & Std. Error & df & $t$ value  & Pr($>t$) & \\ 
\midrule
 (Intercept)            & $3.181$ & $0.057$ & $222.1$  & $55.83$  & $< 2e-16$ & *** \\    
 C(Condition)recommended  & $0.042$ & $0.013$ & $31620$  & $3.29$  & $0.000989$ & *** \\   
\bottomrule
\end{tabular}
\label{rating}
\end{table}

\begin{table}[h]
\centering
\caption{Hyperparemeter search space for the XGBoost model.}
\begin{tabular}{lc}
 Name of the hyperparameter & Search values  \\ 
\midrule
 Number of estimators           & $[100, 200, 300]$ \\   
Learning rate          & $[0.001, 0.01, 0.05, 0.1, 0.2]$ \\  
Maximum depth          & $[3, 5, 7, 10]$ \\  
Subsample         & $[0.5, 0.75, 1]$ \\  
Colsample by tree         & $[0.5, 0.75, 1]$ \\  
Number of leaves & $[25, 30, 35]$\\
\bottomrule
\end{tabular}
\label{hyperparameters}
\end{table}

\clearpage

\begin{table}[h]
\centering
\caption{Test-set performances of different models in predicting marginal distinct idea counts.}
\begin{tabular}{lrr}
Model & $R^2$ & Mean Absolute Error ($MAE$) \\ 
\midrule
Support Vector Regression (SVR)          & $27.71\%$  & 1.55 \\  
Ridge                                    & $30.84\%$  & 1.48 \\ 
LightGBM                                 & $31.11\%$  & 1.47 \\  
RandomForest                             & $31.65\%$  & 1.46 \\  
XGBoost                                  & $\textbf{32.58\%}$  & \textbf{1.44} \\   
\bottomrule
\end{tabular}
\label{model_results}
\end{table}

\clearpage

\end{appendices}


\begin{thebibliography}{82}
\ifx \bisbn   \undefined \def \bisbn  #1{ISBN #1}\fi
\ifx \binits  \undefined \def \binits#1{#1}\fi
\ifx \bauthor  \undefined \def \bauthor#1{#1}\fi
\ifx \batitle  \undefined \def \batitle#1{#1}\fi
\ifx \bjtitle  \undefined \def \bjtitle#1{#1}\fi
\ifx \bvolume  \undefined \def \bvolume#1{\textbf{#1}}\fi
\ifx \byear  \undefined \def \byear#1{#1}\fi
\ifx \bissue  \undefined \def \bissue#1{#1}\fi
\ifx \bfpage  \undefined \def \bfpage#1{#1}\fi
\ifx \blpage  \undefined \def \blpage #1{#1}\fi
\ifx \burl  \undefined \def \burl#1{\textsf{#1}}\fi
\ifx \doiurl  \undefined \def \doiurl#1{\url{https://doi.org/#1}}\fi
\ifx \betal  \undefined \def \betal{\textit{et al.}}\fi
\ifx \binstitute  \undefined \def \binstitute#1{#1}\fi
\ifx \binstitutionaled  \undefined \def \binstitutionaled#1{#1}\fi
\ifx \bctitle  \undefined \def \bctitle#1{#1}\fi
\ifx \beditor  \undefined \def \beditor#1{#1}\fi
\ifx \bpublisher  \undefined \def \bpublisher#1{#1}\fi
\ifx \bbtitle  \undefined \def \bbtitle#1{#1}\fi
\ifx \bedition  \undefined \def \bedition#1{#1}\fi
\ifx \bseriesno  \undefined \def \bseriesno#1{#1}\fi
\ifx \blocation  \undefined \def \blocation#1{#1}\fi
\ifx \bsertitle  \undefined \def \bsertitle#1{#1}\fi
\ifx \bsnm \undefined \def \bsnm#1{#1}\fi
\ifx \bsuffix \undefined \def \bsuffix#1{#1}\fi
\ifx \bparticle \undefined \def \bparticle#1{#1}\fi
\ifx \barticle \undefined \def \barticle#1{#1}\fi
\bibcommenthead
\ifx \bconfdate \undefined \def \bconfdate #1{#1}\fi
\ifx \botherref \undefined \def \botherref #1{#1}\fi
\ifx \url \undefined \def \url#1{\textsf{#1}}\fi
\ifx \bchapter \undefined \def \bchapter#1{#1}\fi
\ifx \bbook \undefined \def \bbook#1{#1}\fi
\ifx \bcomment \undefined \def \bcomment#1{#1}\fi
\ifx \oauthor \undefined \def \oauthor#1{#1}\fi
\ifx \citeauthoryear \undefined \def \citeauthoryear#1{#1}\fi
\ifx \endbibitem  \undefined \def \endbibitem {}\fi
\ifx \bconflocation  \undefined \def \bconflocation#1{#1}\fi
\ifx \arxivurl  \undefined \def \arxivurl#1{\textsf{#1}}\fi
\csname PreBibitemsHook\endcsname

\bibitem[\protect\citeauthoryear{Hofstra et~al.}{2020}]{hofstra2020diversity}
\begin{barticle}
\bauthor{\bsnm{Hofstra}, \binits{B.}},
\bauthor{\bsnm{Kulkarni}, \binits{V.V.}},
\bauthor{\bsnm{Galvez}, \binits{S.M.-N.}},
\bauthor{\bsnm{He}, \binits{B.}},
\bauthor{\bsnm{Jurafsky}, \binits{D.}},
\bauthor{\bsnm{McFarland}, \binits{D.A.}}:
\batitle{The diversity--innovation paradox in science}.
\bjtitle{Proceedings of the National Academy of Sciences}
\bvolume{117}(\bissue{17}),
\bfpage{9284}--\blpage{9291}
(\byear{2020})
\end{barticle}
\endbibitem

\bibitem[\protect\citeauthoryear{Uzzi et~al.}{2013}]{uzzi2013atypical}
\begin{barticle}
\bauthor{\bsnm{Uzzi}, \binits{B.}},
\bauthor{\bsnm{Mukherjee}, \binits{S.}},
\bauthor{\bsnm{Stringer}, \binits{M.}},
\bauthor{\bsnm{Jones}, \binits{B.}}:
\batitle{Atypical combinations and scientific impact}.
\bjtitle{Science}
\bvolume{342}(\bissue{6157}),
\bfpage{468}--\blpage{472}
(\byear{2013})
\end{barticle}
\endbibitem

\bibitem[\protect\citeauthoryear{Li et~al.}{2020}]{li2020attribute}
\begin{barticle}
\bauthor{\bsnm{Li}, \binits{J.}},
\bauthor{\bsnm{Yang}, \binits{J.}},
\bauthor{\bsnm{Zhang}, \binits{J.}},
\bauthor{\bsnm{Liu}, \binits{C.}},
\bauthor{\bsnm{Wang}, \binits{C.}},
\bauthor{\bsnm{Xu}, \binits{T.}}:
\batitle{Attribute-conditioned layout {GAN} for automatic graphic design}.
\bjtitle{IEEE Transactions on Visualization and Computer Graphics}
\bvolume{27}(\bissue{10}),
\bfpage{4039}--\blpage{4048}
(\byear{2020})
\end{barticle}
\endbibitem

\bibitem[\protect\citeauthoryear{Jia et~al.}{2024}]{jia2024simulbench}
\begin{botherref}
\oauthor{\bsnm{Jia}, \binits{Q.}},
\oauthor{\bsnm{Yue}, \binits{X.}},
\oauthor{\bsnm{Zheng}, \binits{T.}},
\oauthor{\bsnm{Huang}, \binits{J.}},
\oauthor{\bsnm{Lin}, \binits{B.Y.}}:
Simul{B}ench: Evaluating language models with creative simulation tasks.
arXiv preprint arXiv:2409.07641
(2024)
\end{botherref}
\endbibitem

\bibitem[\protect\citeauthoryear{Davis et~al.}{2013}]{davis2013toward}
\begin{bchapter}
\bauthor{\bsnm{Davis}, \binits{N.}},
\bauthor{\bsnm{Winnem{\"o}ller}, \binits{H.}},
\bauthor{\bsnm{Dontcheva}, \binits{M.}},
\bauthor{\bsnm{Do}, \binits{E.Y.-L.}}:
\bctitle{Toward a cognitive theory of creativity support}.
In: \bbtitle{Proceedings of the 9th ACM Conference on Creativity \& Cognition},
pp. \bfpage{13}--\blpage{22}
(\byear{2013})
\end{bchapter}
\endbibitem

\bibitem[\protect\citeauthoryear{Jeon et~al.}{2021}]{jeon2021fashionq}
\begin{bchapter}
\bauthor{\bsnm{Jeon}, \binits{Y.}},
\bauthor{\bsnm{Jin}, \binits{S.}},
\bauthor{\bsnm{Shih}, \binits{P.C.}},
\bauthor{\bsnm{Han}, \binits{K.}}:
\bctitle{Fashionq: an {AI}-driven creativity support tool for facilitating ideation in fashion design}.
In: \bbtitle{Proceedings of the 2021 CHI Conference on Human Factors in Computing Systems},
pp. \bfpage{1}--\blpage{18}
(\byear{2021})
\end{bchapter}
\endbibitem

\bibitem[\protect\citeauthoryear{Shih et~al.}{2011}]{shih2011brainstorming}
\begin{bchapter}
\bauthor{\bsnm{Shih}, \binits{P.C.}},
\bauthor{\bsnm{Venolia}, \binits{G.}},
\bauthor{\bsnm{Olson}, \binits{G.M.}}:
\bctitle{Brainstorming under constraints: Why software developers brainstorm in groups}.
In: \bbtitle{Proceedings of the 25th BCS Conference on Human-Computer Interaction},
pp. \bfpage{74}--\blpage{83}
(\byear{2011})
\end{bchapter}
\endbibitem

\bibitem[\protect\citeauthoryear{Peters et~al.}{2024}]{peters2024context}
\begin{barticle}
\bauthor{\bsnm{Peters}, \binits{H.}},
\bauthor{\bsnm{Liu}, \binits{Y.}},
\bauthor{\bsnm{Barbieri}, \binits{F.}},
\bauthor{\bsnm{Baten}, \binits{R.A.}},
\bauthor{\bsnm{Matz}, \binits{S.C.}},
\bauthor{\bsnm{Bos}, \binits{M.W.}}:
\batitle{Context-aware prediction of active and passive user engagement: Evidence from a large online social platform}.
\bjtitle{Journal of Big Data}
\bvolume{11}(\bissue{1}),
\bfpage{110}
(\byear{2024})
\end{barticle}
\endbibitem

\bibitem[\protect\citeauthoryear{Liu et~al.}{2019}]{liu2019characterizing}
\begin{bchapter}
\bauthor{\bsnm{Liu}, \binits{Y.}},
\bauthor{\bsnm{Shi}, \binits{X.}},
\bauthor{\bsnm{Pierce}, \binits{L.}},
\bauthor{\bsnm{Ren}, \binits{X.}}:
\bctitle{Characterizing and forecasting user engagement with in-app action graph: A case study of {S}napchat}.
In: \bbtitle{Proceedings of the 25th ACM SIGKDD International Conference on Knowledge Discovery \& Data Mining},
pp. \bfpage{2023}--\blpage{2031}
(\byear{2019})
\end{bchapter}
\endbibitem

\bibitem[\protect\citeauthoryear{Xia et~al.}{2023}]{xia2023deep}
\begin{bchapter}
\bauthor{\bsnm{Xia}, \binits{Y.}},
\bauthor{\bsnm{Cao}, \binits{Y.}},
\bauthor{\bsnm{Hu}, \binits{S.}},
\bauthor{\bsnm{Liu}, \binits{T.}},
\bauthor{\bsnm{Lu}, \binits{L.}}:
\bctitle{Deep intention-aware network for click-through rate prediction}.
In: \bbtitle{Companion Proceedings of the ACM Web Conference 2023},
pp. \bfpage{533}--\blpage{537}
(\byear{2023})
\end{bchapter}
\endbibitem

\bibitem[\protect\citeauthoryear{Perc and Szolnoki}{2010}]{perc2010coevolutionary}
\begin{barticle}
\bauthor{\bsnm{Perc}, \binits{M.}},
\bauthor{\bsnm{Szolnoki}, \binits{A.}}:
\batitle{Coevolutionary games—a mini review}.
\bjtitle{BioSystems}
\bvolume{99}(\bissue{2}),
\bfpage{109}--\blpage{125}
(\byear{2010})
\end{barticle}
\endbibitem

\bibitem[\protect\citeauthoryear{Kelty et~al.}{2023}]{kelty2023don}
\begin{botherref}
\oauthor{\bsnm{Kelty}, \binits{S.}},
\oauthor{\bsnm{Baten}, \binits{R.A.}},
\oauthor{\bsnm{Proma}, \binits{A.M.}},
\oauthor{\bsnm{Hoque}, \binits{E.}},
\oauthor{\bsnm{Bollen}, \binits{J.}},
\oauthor{\bsnm{Ghoshal}, \binits{G.}}:
Don't follow the leader: Independent thinkers create scientific innovation.
arXiv preprint arXiv:2301.02396
(2023)
\end{botherref}
\endbibitem

\bibitem[\protect\citeauthoryear{Henrich}{2016}]{henrich2016secret}
\begin{bbook}
\bauthor{\bsnm{Henrich}, \binits{J.}}:
\bbtitle{The Secret of Our Success: How Culture Is Driving Human Evolution, Domesticating Our Species, and Making Us Smarter}.
\bpublisher{Princeton University Press},
\blocation{Princeton, New Jersey}
(\byear{2016}).
\doiurl{10.1515/9781400873296}
\end{bbook}
\endbibitem

\bibitem[\protect\citeauthoryear{Herrmann et~al.}{2007}]{herrmann2007humans}
\begin{barticle}
\bauthor{\bsnm{Herrmann}, \binits{E.}},
\bauthor{\bsnm{Call}, \binits{J.}},
\bauthor{\bsnm{Hern{\'a}ndez-Lloreda}, \binits{M.V.}},
\bauthor{\bsnm{Hare}, \binits{B.}},
\bauthor{\bsnm{Tomasello}, \binits{M.}}:
\batitle{Humans have evolved specialized skills of social cognition: The cultural intelligence hypothesis}.
\bjtitle{Science}
\bvolume{317}(\bissue{5843}),
\bfpage{1360}--\blpage{1366}
(\byear{2007})
\end{barticle}
\endbibitem

\bibitem[\protect\citeauthoryear{Boyd et~al.}{2011}]{boyd2011cultural}
\begin{barticle}
\bauthor{\bsnm{Boyd}, \binits{R.}},
\bauthor{\bsnm{Richerson}, \binits{P.J.}},
\bauthor{\bsnm{Henrich}, \binits{J.}}:
\batitle{The cultural niche: Why social learning is essential for human adaptation}.
\bjtitle{Proceedings of the National Academy of Sciences}
\bvolume{108}(\bissue{Supplement 2}),
\bfpage{10918}--\blpage{10925}
(\byear{2011})
\end{barticle}
\endbibitem

\bibitem[\protect\citeauthoryear{Rand et~al.}{2011}]{rand2011dynamic}
\begin{barticle}
\bauthor{\bsnm{Rand}, \binits{D.G.}},
\bauthor{\bsnm{Arbesman}, \binits{S.}},
\bauthor{\bsnm{Christakis}, \binits{N.A.}}:
\batitle{Dynamic social networks promote cooperation in experiments with humans}.
\bjtitle{Proceedings of the National Academy of Sciences}
\bvolume{108}(\bissue{48}),
\bfpage{19193}--\blpage{19198}
(\byear{2011})
\end{barticle}
\endbibitem

\bibitem[\protect\citeauthoryear{Szolnoki et~al.}{2008}]{szolnoki2008making}
\begin{barticle}
\bauthor{\bsnm{Szolnoki}, \binits{A.}},
\bauthor{\bsnm{Perc}, \binits{M.}},
\bauthor{\bsnm{Danku}, \binits{Z.}}:
\batitle{Making new connections towards cooperation in the {P}risoner's {D}ilemma game}.
\bjtitle{EPL (Europhysics Letters)}
\bvolume{84}(\bissue{5}),
\bfpage{50007}
(\byear{2008})
\end{barticle}
\endbibitem

\bibitem[\protect\citeauthoryear{Bernstein et~al.}{2018}]{bernstein2018intermittent}
\begin{barticle}
\bauthor{\bsnm{Bernstein}, \binits{E.}},
\bauthor{\bsnm{Shore}, \binits{J.}},
\bauthor{\bsnm{Lazer}, \binits{D.}}:
\batitle{How intermittent breaks in interaction improve collective intelligence}.
\bjtitle{Proceedings of the National Academy of Sciences}
\bvolume{115}(\bissue{35}),
\bfpage{8734}--\blpage{8739}
(\byear{2018})
\end{barticle}
\endbibitem

\bibitem[\protect\citeauthoryear{Almaatouq et~al.}{2020}]{almaatouq2020adaptive}
\begin{barticle}
\bauthor{\bsnm{Almaatouq}, \binits{A.}},
\bauthor{\bsnm{Noriega-Campero}, \binits{A.}},
\bauthor{\bsnm{Alotaibi}, \binits{A.}},
\bauthor{\bsnm{Krafft}, \binits{P.M.}},
\bauthor{\bsnm{Moussaid}, \binits{M.}},
\bauthor{\bsnm{Pentland}, \binits{A.}}:
\batitle{Adaptive social networks promote the wisdom of crowds}.
\bjtitle{Proceedings of the National Academy of Sciences}
\bvolume{117}(\bissue{21}),
\bfpage{11379}--\blpage{11386}
(\byear{2020})
\doiurl{10.1073/pnas.1917687117}
\end{barticle}
\endbibitem

\bibitem[\protect\citeauthoryear{Shafipour et~al.}{2018}]{shafipour2018buildup}
\begin{barticle}
\bauthor{\bsnm{Shafipour}, \binits{R.}},
\bauthor{\bsnm{Baten}, \binits{R.A.}},
\bauthor{\bsnm{Hasan}, \binits{M.K.}},
\bauthor{\bsnm{Ghoshal}, \binits{G.}},
\bauthor{\bsnm{Mateos}, \binits{G.}},
\bauthor{\bsnm{Hoque}, \binits{M.E.}}:
\batitle{Buildup of speaking skills in an online learning community: A network-analytic exploration}.
\bjtitle{Palgrave Communications}
\bvolume{4}(\bissue{1}),
\bfpage{63}
(\byear{2018})
\end{barticle}
\endbibitem

\bibitem[\protect\citeauthoryear{Baten et~al.}{2019}]{baten2019upskilling}
\begin{bchapter}
\bauthor{\bsnm{Baten}, \binits{R.A.}},
\bauthor{\bsnm{Clark}, \binits{F.}},
\bauthor{\bsnm{Hoque}, \binits{M.E.}}:
\bctitle{Upskilling together: How peer-interaction influences speaking-skills development online}.
In: \bbtitle{8th International Conference on Affective Computing and Intelligent Interaction (ACII)},
pp. \bfpage{662}--\blpage{668}
(\byear{2019}).
\bcomment{IEEE}
\end{bchapter}
\endbibitem

\bibitem[\protect\citeauthoryear{Baten et~al.}{2020}]{baten2020creativity}
\begin{barticle}
\bauthor{\bsnm{Baten}, \binits{R.A.}},
\bauthor{\bsnm{Bagley}, \binits{D.}},
\bauthor{\bsnm{Tenesaca}, \binits{A.}},
\bauthor{\bsnm{Clark}, \binits{F.}},
\bauthor{\bsnm{Bagrow}, \binits{J.P.}},
\bauthor{\bsnm{Ghoshal}, \binits{G.}},
\bauthor{\bsnm{Hoque}, \binits{E.}}:
\batitle{Creativity in temporal social networks: How divergent thinking is impacted by one’s choice of peers}.
\bjtitle{Journal of the Royal Society Interface}
\bvolume{17}(\bissue{171}),
\bfpage{20200667}
(\byear{2020})
\end{barticle}
\endbibitem

\bibitem[\protect\citeauthoryear{Baten et~al.}{2021}]{baten2021cues}
\begin{barticle}
\bauthor{\bsnm{Baten}, \binits{R.A.}},
\bauthor{\bsnm{Aslin}, \binits{R.N.}},
\bauthor{\bsnm{Ghoshal}, \binits{G.}},
\bauthor{\bsnm{Hoque}, \binits{E.}}:
\batitle{Cues to gender and racial identity reduce creativity in diverse social networks}.
\bjtitle{Scientific Reports}
\bvolume{11}(\bissue{1}),
\bfpage{1}--\blpage{10}
(\byear{2021})
\end{barticle}
\endbibitem

\bibitem[\protect\citeauthoryear{Baten et~al.}{2022}]{baten2022novel}
\begin{barticle}
\bauthor{\bsnm{Baten}, \binits{R.A.}},
\bauthor{\bsnm{Aslin}, \binits{R.N.}},
\bauthor{\bsnm{Ghoshal}, \binits{G.}},
\bauthor{\bsnm{Hoque}, \binits{E.}}:
\batitle{Novel idea generation in social networks is optimized by exposure to a ``{G}oldilocks'' level of idea-variability}.
\bjtitle{PNAS Nexus}
\bvolume{1}(\bissue{5}),
\bfpage{255}
(\byear{2022})
\end{barticle}
\endbibitem

\bibitem[\protect\citeauthoryear{Guilford et~al.}{1978}]{guildford1978alternate}
\begin{botherref}
\oauthor{\bsnm{Guilford}, \binits{J.}},
\oauthor{\bsnm{Christensen}, \binits{P.}},
\oauthor{\bsnm{Merrifield}, \binits{P.}},
\oauthor{\bsnm{Wilson}, \binits{R.}}:
Alternate Uses: Manual of Instructions and Interpretation.
Orange, CA: Sheridan Psychological Services,
(1978).
Orange, CA: Sheridan Psychological Services
\end{botherref}
\endbibitem

\bibitem[\protect\citeauthoryear{Kenett}{2023}]{kenett2023creatively}
\begin{botherref}
\oauthor{\bsnm{Kenett}, \binits{Y.N.}}:
Creatively searching through semantic memory structure: A short integrative review.
The Routledge International Handbook of Creative Cognition,
160--179
(2023)
\end{botherref}
\endbibitem

\bibitem[\protect\citeauthoryear{Brown and Paulus}{2002}]{brown2002making}
\begin{barticle}
\bauthor{\bsnm{Brown}, \binits{V.R.}},
\bauthor{\bsnm{Paulus}, \binits{P.B.}}:
\batitle{Making group brainstorming more effective: Recommendations from an associative memory perspective}.
\bjtitle{Current Directions in Psychological Science}
\bvolume{11}(\bissue{6}),
\bfpage{208}--\blpage{212}
(\byear{2002})
\end{barticle}
\endbibitem

\bibitem[\protect\citeauthoryear{Mednick}{1962}]{mednick1962associative}
\begin{barticle}
\bauthor{\bsnm{Mednick}, \binits{S.}}:
\batitle{The associative basis of the creative process}.
\bjtitle{Psychological Review}
\bvolume{69}(\bissue{3}),
\bfpage{220}
(\byear{1962})
\end{barticle}
\endbibitem

\bibitem[\protect\citeauthoryear{Nijstad and Stroebe}{2006}]{nijstad2006group}
\begin{barticle}
\bauthor{\bsnm{Nijstad}, \binits{B.A.}},
\bauthor{\bsnm{Stroebe}, \binits{W.}}:
\batitle{How the group affects the mind: A cognitive model of idea generation in groups}.
\bjtitle{Personality and Social Psychology Review}
\bvolume{10}(\bissue{3}),
\bfpage{186}--\blpage{213}
(\byear{2006})
\end{barticle}
\endbibitem

\bibitem[\protect\citeauthoryear{Runco}{2014}]{runco2014creativity}
\begin{bbook}
\bauthor{\bsnm{Runco}, \binits{M.A.}}:
\bbtitle{Creativity: Theories and Themes: Research, Development, and Practice}.
\bpublisher{Elsevier}, \blocation{???}
(\byear{2014})
\end{bbook}
\endbibitem

\bibitem[\protect\citeauthoryear{Lundberg and Lee}{2017}]{scott2017unified}
\begin{bchapter}
\bauthor{\bsnm{Lundberg}, \binits{S.M.}},
\bauthor{\bsnm{Lee}, \binits{S.-I.}}:
\bctitle{A unified approach to interpreting model predictions}.
In: \beditor{\bsnm{Guyon}, \binits{I.}},
\beditor{\bsnm{Luxburg}, \binits{U.V.}},
\beditor{\bsnm{Bengio}, \binits{S.}},
\beditor{\bsnm{Wallach}, \binits{H.}},
\beditor{\bsnm{Fergus}, \binits{R.}},
\beditor{\bsnm{Vishwanathan}, \binits{S.}},
\beditor{\bsnm{Garnett}, \binits{R.}} (eds.)
\bbtitle{Advances in Neural Information Processing Systems},
vol. \bseriesno{30},
pp. \bfpage{4765}--\blpage{4774}.
\bpublisher{Curran Associates, Inc.}, \blocation{???}
(\byear{2017})
\end{bchapter}
\endbibitem

\bibitem[\protect\citeauthoryear{Beaty and Johnson}{2021}]{beaty2021automating}
\begin{barticle}
\bauthor{\bsnm{Beaty}, \binits{R.E.}},
\bauthor{\bsnm{Johnson}, \binits{D.R.}}:
\batitle{Automating creativity assessment with {S}em{D}is: An open platform for computing semantic distance}.
\bjtitle{Behavior Research Methods}
\bvolume{53}(\bissue{2}),
\bfpage{757}--\blpage{780}
(\byear{2021})
\end{barticle}
\endbibitem

\bibitem[\protect\citeauthoryear{Lundberg et~al.}{2020}]{lundberg2020local}
\begin{barticle}
\bauthor{\bsnm{Lundberg}, \binits{S.M.}},
\bauthor{\bsnm{Erion}, \binits{G.}},
\bauthor{\bsnm{Chen}, \binits{H.}},
\bauthor{\bsnm{DeGrave}, \binits{A.}},
\bauthor{\bsnm{Prutkin}, \binits{J.M.}},
\bauthor{\bsnm{Nair}, \binits{B.}},
\bauthor{\bsnm{Katz}, \binits{R.}},
\bauthor{\bsnm{Himmelfarb}, \binits{J.}},
\bauthor{\bsnm{Bansal}, \binits{N.}},
\bauthor{\bsnm{Lee}, \binits{S.-I.}}:
\batitle{From local explanations to global understanding with explainable {AI} for trees}.
\bjtitle{Nature Machine Intelligence}
\bvolume{2}(\bissue{1}),
\bfpage{56}--\blpage{67}
(\byear{2020})
\end{barticle}
\endbibitem

\bibitem[\protect\citeauthoryear{Molden}{2014}]{molden2014understanding}
\begin{barticle}
\bauthor{\bsnm{Molden}, \binits{D.C.}}:
\batitle{Understanding priming effects in social psychology: What is ``social priming'' and how does it occur?}
\bjtitle{Social Cognition}
\bvolume{32}(\bissue{Supplement}),
\bfpage{1}--\blpage{11}
(\byear{2014})
\end{barticle}
\endbibitem

\bibitem[\protect\citeauthoryear{Kahneman}{2011}]{kahneman2011thinking}
\begin{bbook}
\bauthor{\bsnm{Kahneman}, \binits{D.}}:
\bbtitle{Thinking, Fast and Slow}.
\bpublisher{Farrar, Straus and Giroux},
\blocation{New York, NY, USA}
(\byear{2011})
\end{bbook}
\endbibitem

\bibitem[\protect\citeauthoryear{Acar et~al.}{2023}]{acar2023collective}
\begin{botherref}
\oauthor{\bsnm{Acar}, \binits{O.A.}},
\oauthor{\bsnm{Tuncdogan}, \binits{A.}},
\oauthor{\bsnm{Knippenberg}, \binits{D.}},
\oauthor{\bsnm{Lakhani}, \binits{K.R.}}:
Collective creativity and innovation: An interdisciplinary review, integration, and research agenda.
Journal of Management,
01492063231212416
(2023)
\end{botherref}
\endbibitem

\bibitem[\protect\citeauthoryear{Perry-Smith and Shalley}{2003}]{perry2003social}
\begin{barticle}
\bauthor{\bsnm{Perry-Smith}, \binits{J.E.}},
\bauthor{\bsnm{Shalley}, \binits{C.E.}}:
\batitle{The social side of creativity: A static and dynamic social network perspective}.
\bjtitle{Academy of Management Review}
\bvolume{28}(\bissue{1}),
\bfpage{89}--\blpage{106}
(\byear{2003})
\end{barticle}
\endbibitem

\bibitem[\protect\citeauthoryear{Jiang et~al.}{2018}]{jiang2018dynamic}
\begin{barticle}
\bauthor{\bsnm{Jiang}, \binits{H.}},
\bauthor{\bsnm{Zhang}, \binits{Q.-P.}},
\bauthor{\bsnm{Zhou}, \binits{Y.}}:
\batitle{Dynamic creative interaction networks and team creativity evolution: A longitudinal study}.
\bjtitle{The Journal of Creative Behavior}
\bvolume{52}(\bissue{2}),
\bfpage{168}--\blpage{196}
(\byear{2018})
\end{barticle}
\endbibitem

\bibitem[\protect\citeauthoryear{Hossain}{2021}]{hossain2021fast}
\begin{botherref}
\oauthor{\bsnm{Hossain}, \binits{R.}}:
Fast and secure friend recommendation in online social networks.
Master's thesis,
Montclair State University
(2021)
\end{botherref}
\endbibitem

\bibitem[\protect\citeauthoryear{Ning et~al.}{2019}]{Ning2019personet}
\begin{barticle}
\bauthor{\bsnm{Ning}, \binits{H.}},
\bauthor{\bsnm{Dhelim}, \binits{S.}},
\bauthor{\bsnm{Aung}, \binits{N.}}:
\batitle{Personet: Friend recommendation system based on big-five personality traits and hybrid filtering}.
\bjtitle{IEEE Transactions on Computational Social Systems}
\bvolume{6}(\bissue{3}),
\bfpage{394}--\blpage{402}
(\byear{2019})
\doiurl{10.1109/TCSS.2019.2903857}
\end{barticle}
\endbibitem

\bibitem[\protect\citeauthoryear{Cheng et~al.}{2019}]{cheng2019friend}
\begin{barticle}
\bauthor{\bsnm{Cheng}, \binits{S.}},
\bauthor{\bsnm{Zhang}, \binits{B.}},
\bauthor{\bsnm{Zou}, \binits{G.}},
\bauthor{\bsnm{Huang}, \binits{M.}},
\bauthor{\bsnm{Zhang}, \binits{Z.}}:
\batitle{Friend recommendation in social networks based on multi-source information fusion}.
\bjtitle{International Journal of Machine Learning and Cybernetics}
\bvolume{10}(\bissue{5}),
\bfpage{1003}--\blpage{1024}
(\byear{2019})
\end{barticle}
\endbibitem

\bibitem[\protect\citeauthoryear{Proma et~al.}{2024}]{proma2024exploring}
\begin{botherref}
\oauthor{\bsnm{Proma}, \binits{A.M.}},
\oauthor{\bsnm{Pate}, \binits{N.}},
\oauthor{\bsnm{Baten}, \binits{R.A.}},
\oauthor{\bsnm{Chen}, \binits{S.}},
\oauthor{\bsnm{Druckman}, \binits{J.}},
\oauthor{\bsnm{Ghoshal}, \binits{G.}},
\oauthor{\bsnm{Hoque}, \binits{E.}}:
Exploring the role of randomization on belief rigidity in online social networks.
arXiv preprint arXiv:2407.01820
(2024)
\end{botherref}
\endbibitem

\bibitem[\protect\citeauthoryear{He et~al.}{2016}]{he2016vista}
\begin{bchapter}
\bauthor{\bsnm{He}, \binits{R.}},
\bauthor{\bsnm{Fang}, \binits{C.}},
\bauthor{\bsnm{Wang}, \binits{Z.}},
\bauthor{\bsnm{McAuley}, \binits{J.}}:
\bctitle{Vista: A visually, socially, and temporally-aware model for artistic recommendation}.
In: \bbtitle{Proceedings of the 10th ACM Conference on Recommender Systems},
pp. \bfpage{309}--\blpage{316}
(\byear{2016})
\end{bchapter}
\endbibitem

\bibitem[\protect\citeauthoryear{Ziegler et~al.}{2005}]{ziegler2005improving}
\begin{bchapter}
\bauthor{\bsnm{Ziegler}, \binits{C.-N.}},
\bauthor{\bsnm{McNee}, \binits{S.M.}},
\bauthor{\bsnm{Konstan}, \binits{J.A.}},
\bauthor{\bsnm{Lausen}, \binits{G.}}:
\bctitle{Improving recommendation lists through topic diversification}.
In: \bbtitle{Proceedings of the 14th International Conference on World Wide Web},
pp. \bfpage{22}--\blpage{32}
(\byear{2005})
\end{bchapter}
\endbibitem

\bibitem[\protect\citeauthoryear{Kamath et~al.}{2013}]{kamath2013board}
\begin{bchapter}
\bauthor{\bsnm{Kamath}, \binits{K.Y.}},
\bauthor{\bsnm{Popescu}, \binits{A.-M.}},
\bauthor{\bsnm{Caverlee}, \binits{J.}}:
\bctitle{Board recommendation in {P}interest}.
In: \bbtitle{UMAP Workshops},
p. \bfpage{28}
(\byear{2013}).
\bcomment{Citeseer}
\end{bchapter}
\endbibitem

\bibitem[\protect\citeauthoryear{Kislyuk et~al.}{2015}]{kislyuk2015human}
\begin{botherref}
\oauthor{\bsnm{Kislyuk}, \binits{D.}},
\oauthor{\bsnm{Liu}, \binits{Y.}},
\oauthor{\bsnm{Liu}, \binits{D.}},
\oauthor{\bsnm{Tzeng}, \binits{E.}},
\oauthor{\bsnm{Jing}, \binits{Y.}}:
Human curation and {C}onv{N}ets: Powering item-to-item recommendations on {P}interest.
arXiv preprint arXiv:1511.04003
(2015)
\end{botherref}
\endbibitem

\bibitem[\protect\citeauthoryear{Chen and Guestrin}{2016}]{chen2016xgboost}
\begin{bchapter}
\bauthor{\bsnm{Chen}, \binits{T.}},
\bauthor{\bsnm{Guestrin}, \binits{C.}}:
\bctitle{{XGB}oost: A scalable tree boosting system}.
In: \bbtitle{Proceedings of the 22nd ACM SIGKDD International Conference on Knowledge Discovery and Data Mining},
pp. \bfpage{785}--\blpage{794}
(\byear{2016})
\end{bchapter}
\endbibitem

\bibitem[\protect\citeauthoryear{Han et~al.}{2012}]{han2012data}
\begin{bbook}
\bauthor{\bsnm{Han}, \binits{J.}},
\bauthor{\bsnm{Kamber}, \binits{M.}},
\bauthor{\bsnm{Pei}, \binits{J.}}:
\bbtitle{Data Mining: Concepts and Techniques}.
\bpublisher{Waltham: Morgan Kaufmann Publishers}, \blocation{???}
(\byear{2012})
\end{bbook}
\endbibitem

\bibitem[\protect\citeauthoryear{Baten et~al.}{2023}]{baten2023predicting}
\begin{bchapter}
\bauthor{\bsnm{Baten}, \binits{R.A.}},
\bauthor{\bsnm{Liu}, \binits{Y.}},
\bauthor{\bsnm{Peters}, \binits{H.}},
\bauthor{\bsnm{Barbieri}, \binits{F.}},
\bauthor{\bsnm{Shah}, \binits{N.}},
\bauthor{\bsnm{Neves}, \binits{L.}},
\bauthor{\bsnm{Bos}, \binits{M.W.}}:
\bctitle{Predicting future location categories of users in a large social platform}.
In: \bbtitle{Proceedings of the International AAAI Conference on Web and Social Media},
vol. \bseriesno{17},
pp. \bfpage{47}--\blpage{58}
(\byear{2023})
\end{bchapter}
\endbibitem

\bibitem[\protect\citeauthoryear{Covington et~al.}{2016}]{covington2016deep}
\begin{bchapter}
\bauthor{\bsnm{Covington}, \binits{P.}},
\bauthor{\bsnm{Adams}, \binits{J.}},
\bauthor{\bsnm{Sargin}, \binits{E.}}:
\bctitle{Deep neural networks for {Y}ou{T}ube recommendations}.
In: \bbtitle{Proceedings of the 10th ACM Conference on Recommender Systems},
pp. \bfpage{191}--\blpage{198}
(\byear{2016})
\end{bchapter}
\endbibitem

\bibitem[\protect\citeauthoryear{Kolodner et~al.}{2024}]{kolodner2024robust}
\begin{botherref}
\oauthor{\bsnm{Kolodner}, \binits{M.}},
\oauthor{\bsnm{Ju}, \binits{M.}},
\oauthor{\bsnm{Fan}, \binits{Z.}},
\oauthor{\bsnm{Zhao}, \binits{T.}},
\oauthor{\bsnm{Ghazizadeh}, \binits{E.}},
\oauthor{\bsnm{Wu}, \binits{Y.}},
\oauthor{\bsnm{Shah}, \binits{N.}},
\oauthor{\bsnm{Liu}, \binits{Y.}}:
Robust training objectives improve embedding-based retrieval in industrial recommendation systems.
arXiv preprint arXiv:2409.14682
(2024)
\end{botherref}
\endbibitem

\bibitem[\protect\citeauthoryear{Azoulay et~al.}{2010}]{azoulay2010superstar}
\begin{barticle}
\bauthor{\bsnm{Azoulay}, \binits{P.}},
\bauthor{\bsnm{Graff~Zivin}, \binits{J.S.}},
\bauthor{\bsnm{Wang}, \binits{J.}}:
\batitle{Superstar extinction}.
\bjtitle{The Quarterly Journal of Economics}
\bvolume{125}(\bissue{2}),
\bfpage{549}--\blpage{589}
(\byear{2010})
\end{barticle}
\endbibitem

\bibitem[\protect\citeauthoryear{Salganik et~al.}{2006}]{salganik2006experimental}
\begin{barticle}
\bauthor{\bsnm{Salganik}, \binits{M.J.}},
\bauthor{\bsnm{Dodds}, \binits{P.S.}},
\bauthor{\bsnm{Watts}, \binits{D.J.}}:
\batitle{Experimental study of inequality and unpredictability in an artificial cultural market}.
\bjtitle{Science}
\bvolume{311}(\bissue{5762}),
\bfpage{854}--\blpage{856}
(\byear{2006})
\end{barticle}
\endbibitem

\bibitem[\protect\citeauthoryear{Dugosh and Paulus}{2005}]{dugosh2005cognitive}
\begin{barticle}
\bauthor{\bsnm{Dugosh}, \binits{K.L.}},
\bauthor{\bsnm{Paulus}, \binits{P.B.}}:
\batitle{Cognitive and social comparison processes in brainstorming}.
\bjtitle{Journal of Experimental Social Psychology}
\bvolume{41}(\bissue{3}),
\bfpage{313}--\blpage{320}
(\byear{2005})
\end{barticle}
\endbibitem

\bibitem[\protect\citeauthoryear{Paulus}{2000}]{paulus2000groups}
\begin{barticle}
\bauthor{\bsnm{Paulus}, \binits{P.}}:
\batitle{Groups, teams, and creativity: The creative potential of idea-generating groups}.
\bjtitle{Applied Psychology}
\bvolume{49}(\bissue{2}),
\bfpage{237}--\blpage{262}
(\byear{2000})
\end{barticle}
\endbibitem

\bibitem[\protect\citeauthoryear{Brown et~al.}{1998}]{brown1998modeling}
\begin{barticle}
\bauthor{\bsnm{Brown}, \binits{V.}},
\bauthor{\bsnm{Tumeo}, \binits{M.}},
\bauthor{\bsnm{Larey}, \binits{T.S.}},
\bauthor{\bsnm{Paulus}, \binits{P.B.}}:
\batitle{Modeling cognitive interactions during group brainstorming}.
\bjtitle{Small Group Research}
\bvolume{29}(\bissue{4}),
\bfpage{495}--\blpage{526}
(\byear{1998})
\end{barticle}
\endbibitem

\bibitem[\protect\citeauthoryear{Bechtoldt et~al.}{2010}]{bechtoldt2010motivated}
\begin{barticle}
\bauthor{\bsnm{Bechtoldt}, \binits{M.N.}},
\bauthor{\bsnm{De~Dreu}, \binits{C.K.}},
\bauthor{\bsnm{Nijstad}, \binits{B.A.}},
\bauthor{\bsnm{Choi}, \binits{H.-S.}}:
\batitle{Motivated information processing, social tuning, and group creativity}.
\bjtitle{Journal of Personality and Social Psychology}
\bvolume{99}(\bissue{4}),
\bfpage{622}
(\byear{2010})
\end{barticle}
\endbibitem

\bibitem[\protect\citeauthoryear{Scholten et~al.}{2007}]{scholten2007motivated}
\begin{barticle}
\bauthor{\bsnm{Scholten}, \binits{L.}},
\bauthor{\bsnm{Van~Knippenberg}, \binits{D.}},
\bauthor{\bsnm{Nijstad}, \binits{B.A.}},
\bauthor{\bsnm{De~Dreu}, \binits{C.K.}}:
\batitle{Motivated information processing and group decision-making: Effects of process accountability on information processing and decision quality}.
\bjtitle{Journal of Experimental Social Psychology}
\bvolume{43}(\bissue{4}),
\bfpage{539}--\blpage{552}
(\byear{2007})
\end{barticle}
\endbibitem

\bibitem[\protect\citeauthoryear{Wingstr{\"o}m et~al.}{2024}]{wingstrom2024redefining}
\begin{barticle}
\bauthor{\bsnm{Wingstr{\"o}m}, \binits{R.}},
\bauthor{\bsnm{Hautala}, \binits{J.}},
\bauthor{\bsnm{Lundman}, \binits{R.}}:
\batitle{Redefining creativity in the era of {AI}? perspectives of computer scientists and new media artists}.
\bjtitle{Creativity Research Journal}
\bvolume{36}(\bissue{2}),
\bfpage{177}--\blpage{193}
(\byear{2024})
\end{barticle}
\endbibitem

\bibitem[\protect\citeauthoryear{Ivanov}{2023}]{ivanov2023dark}
\begin{barticle}
\bauthor{\bsnm{Ivanov}, \binits{S.}}:
\batitle{The dark side of artificial intelligence in higher education}.
\bjtitle{The Service Industries Journal}
\bvolume{43}(\bissue{15-16}),
\bfpage{1055}--\blpage{1082}
(\byear{2023})
\end{barticle}
\endbibitem

\bibitem[\protect\citeauthoryear{Yakura et~al.}{2024}]{yakura2024empirical}
\begin{botherref}
\oauthor{\bsnm{Yakura}, \binits{H.}},
\oauthor{\bsnm{Lopez-Lopez}, \binits{E.}},
\oauthor{\bsnm{Brinkmann}, \binits{L.}},
\oauthor{\bsnm{Serna}, \binits{I.}},
\oauthor{\bsnm{Gupta}, \binits{P.}},
\oauthor{\bsnm{Rahwan}, \binits{I.}}:
Empirical evidence of {L}arge {L}anguage {M}odel's influence on human spoken communication.
arXiv preprint arXiv:2409.01754
(2024)
\end{botherref}
\endbibitem

\bibitem[\protect\citeauthoryear{Dignum}{2019}]{dignum2019responsible}
\begin{bbook}
\bauthor{\bsnm{Dignum}, \binits{V.}}:
\bbtitle{Responsible Artificial Intelligence: How to Develop and Use {AI} in a Responsible Way}
vol. \bseriesno{2156}.
\bpublisher{Springer}, \blocation{???}
(\byear{2019})
\end{bbook}
\endbibitem

\bibitem[\protect\citeauthoryear{Mikolov et~al.}{2013}]{mikolov2013efficient}
\begin{botherref}
\oauthor{\bsnm{Mikolov}, \binits{T.}},
\oauthor{\bsnm{Chen}, \binits{K.}},
\oauthor{\bsnm{Corrado}, \binits{G.}},
\oauthor{\bsnm{Dean}, \binits{J.}}:
Efficient estimation of word representations in vector space.
arXiv preprint arXiv:1301.3781
(2013)
\end{botherref}
\endbibitem

\bibitem[\protect\citeauthoryear{Kusner et~al.}{2015}]{kusner2015word}
\begin{bchapter}
\bauthor{\bsnm{Kusner}, \binits{M.}},
\bauthor{\bsnm{Sun}, \binits{Y.}},
\bauthor{\bsnm{Kolkin}, \binits{N.}},
\bauthor{\bsnm{Weinberger}, \binits{K.}}:
\bctitle{From word embeddings to document distances}.
In: \bbtitle{International Conference on Machine Learning},
pp. \bfpage{957}--\blpage{966}
(\byear{2015})
\end{bchapter}
\endbibitem

\bibitem[\protect\citeauthoryear{Pennington et~al.}{2014}]{pennington2014glove}
\begin{bchapter}
\bauthor{\bsnm{Pennington}, \binits{J.}},
\bauthor{\bsnm{Socher}, \binits{R.}},
\bauthor{\bsnm{Manning}, \binits{C.D.}}:
\bctitle{Glove: Global vectors for word representation}.
In: \bbtitle{Proceedings of the 2014 Conference on Empirical Methods in Natural Language Processing (EMNLP)},
pp. \bfpage{1532}--\blpage{1543}
(\byear{2014})
\end{bchapter}
\endbibitem

\bibitem[\protect\citeauthoryear{Kozbelt et~al.}{2010}]{kozbelt2010theories}
\begin{bchapter}
\bauthor{\bsnm{Kozbelt}, \binits{A.}},
\bauthor{\bsnm{Beghetto}, \binits{R.A.}},
\bauthor{\bsnm{Runco}, \binits{M.A.}}:
\bctitle{Theories of creativity}.
In: \bbtitle{The Cambridge Handbook of Creativity},
pp. \bfpage{20}--\blpage{47}.
\bpublisher{Cambridge University Press}, \blocation{???}
(\byear{2010}).
\doiurl{10.1017/CBO9780511763205.004}
\end{bchapter}
\endbibitem

\bibitem[\protect\citeauthoryear{Oppezzo and Schwartz}{2014}]{oppezzo2014give}
\begin{barticle}
\bauthor{\bsnm{Oppezzo}, \binits{M.}},
\bauthor{\bsnm{Schwartz}, \binits{D.L.}}:
\batitle{Give your ideas some legs: The positive effect of walking on creative thinking.}
\bjtitle{Journal of Experimental Psychology: Learning, Memory, and Cognition}
\bvolume{40}(\bissue{4}),
\bfpage{1142}
(\byear{2014})
\end{barticle}
\endbibitem

\bibitem[\protect\citeauthoryear{Abdullah et~al.}{2016}]{abdullah2016shining}
\begin{bchapter}
\bauthor{\bsnm{Abdullah}, \binits{S.}},
\bauthor{\bsnm{Czerwinski}, \binits{M.}},
\bauthor{\bsnm{Mark}, \binits{G.}},
\bauthor{\bsnm{Johns}, \binits{P.}}:
\bctitle{Shining (blue) light on creative ability}.
In: \bbtitle{Proceedings of the 2016 ACM International Joint Conference on Pervasive and Ubiquitous Computing},
pp. \bfpage{793}--\blpage{804}
(\byear{2016}).
\bcomment{ACM}
\end{bchapter}
\endbibitem

\bibitem[\protect\citeauthoryear{Snyder et~al.}{2004}]{snyder2004creativity}
\begin{barticle}
\bauthor{\bsnm{Snyder}, \binits{A.}},
\bauthor{\bsnm{Mitchell}, \binits{J.}},
\bauthor{\bsnm{Bossomaier}, \binits{T.}},
\bauthor{\bsnm{Pallier}, \binits{G.}}:
\batitle{The {C}reativity {Q}uotient: An objective scoring of ideational fluency}.
\bjtitle{Creativity Research Journal}
\bvolume{16}(\bissue{4}),
\bfpage{415}--\blpage{419}
(\byear{2004})
\end{barticle}
\endbibitem

\bibitem[\protect\citeauthoryear{Bossomaier et~al.}{2009}]{bossomaier2009semantic}
\begin{barticle}
\bauthor{\bsnm{Bossomaier}, \binits{T.}},
\bauthor{\bsnm{Harr{\'e}}, \binits{M.}},
\bauthor{\bsnm{Knittel}, \binits{A.}},
\bauthor{\bsnm{Snyder}, \binits{A.}}:
\batitle{A semantic network approach to the {C}reativity {Q}uotient ({CQ})}.
\bjtitle{Creativity Research Journal}
\bvolume{21}(\bissue{1}),
\bfpage{64}--\blpage{71}
(\byear{2009})
\end{barticle}
\endbibitem

\bibitem[\protect\citeauthoryear{Bouchard~Jr and Hare}{1970}]{bouchard1970size}
\begin{barticle}
\bauthor{\bsnm{Bouchard~Jr}, \binits{T.J.}},
\bauthor{\bsnm{Hare}, \binits{M.}}:
\batitle{Size, performance, and potential in brainstorming groups}.
\bjtitle{Journal of Applied Psychology}
\bvolume{54}(\bissue{1p1}),
\bfpage{51}
(\byear{1970})
\end{barticle}
\endbibitem

\bibitem[\protect\citeauthoryear{Rietzschel et~al.}{2007}]{rietzschel2007personal}
\begin{barticle}
\bauthor{\bsnm{Rietzschel}, \binits{E.F.}},
\bauthor{\bsnm{De~Dreu}, \binits{C.K.}},
\bauthor{\bsnm{Nijstad}, \binits{B.A.}}:
\batitle{Personal need for structure and creative performance: The moderating influence of fear of invalidity}.
\bjtitle{Personality and Social Psychology Bulletin}
\bvolume{33}(\bissue{6}),
\bfpage{855}--\blpage{866}
(\byear{2007})
\end{barticle}
\endbibitem

\bibitem[\protect\citeauthoryear{Miller}{1995}]{miller1995wordnet}
\begin{barticle}
\bauthor{\bsnm{Miller}, \binits{G.A.}}:
\batitle{Word{N}et: A lexical database for {E}nglish}.
\bjtitle{Communications of the ACM}
\bvolume{38}(\bissue{11}),
\bfpage{39}--\blpage{41}
(\byear{1995})
\end{barticle}
\endbibitem

\bibitem[\protect\citeauthoryear{Seco et~al.}{2004}]{seco2004intrinsic}
\begin{bchapter}
\bauthor{\bsnm{Seco}, \binits{N.}},
\bauthor{\bsnm{Veale}, \binits{T.}},
\bauthor{\bsnm{Hayes}, \binits{J.}}:
\bctitle{An intrinsic information content metric for semantic similarity in {W}ord{N}et}.
In: \bbtitle{Proceedings of the 16th Eureopean Conference on Artificial Intelligence, ECAI},
vol. \bseriesno{16},
p. \bfpage{1089}
(\byear{2004})
\end{bchapter}
\endbibitem

\bibitem[\protect\citeauthoryear{Jiang and Conrath}{1998}]{jiang1997semantic}
\begin{bchapter}
\bauthor{\bsnm{Jiang}, \binits{J.J.}},
\bauthor{\bsnm{Conrath}, \binits{D.W.}}:
\bctitle{Semantic similarity based on corpus statistics and lexical taxonomy}.
In: \bbtitle{Proceedings of the International Conference on Research in Computational Linguistics}
(\byear{1998})
\end{bchapter}
\endbibitem

\bibitem[\protect\citeauthoryear{Santanen et~al.}{2000}]{santanen2000cognitive}
\begin{bchapter}
\bauthor{\bsnm{Santanen}, \binits{E.L.}},
\bauthor{\bsnm{Briggs}, \binits{R.O.}},
\bauthor{\bsnm{De~Vreede}, \binits{G.-J.}}:
\bctitle{The cognitive network model of creativity: A new causal model of creativity and a new brainstorming technique}.
In: \bbtitle{Proceedings of the 33rd Annual Hawaii International Conference on System Sciences},
p. \bfpage{10}
(\byear{2000}).
\bcomment{IEEE}
\end{bchapter}
\endbibitem

\bibitem[\protect\citeauthoryear{Patterson et~al.}{2007}]{patterson2007you}
\begin{barticle}
\bauthor{\bsnm{Patterson}, \binits{K.}},
\bauthor{\bsnm{Nestor}, \binits{P.J.}},
\bauthor{\bsnm{Rogers}, \binits{T.T.}}:
\batitle{Where do you know what you know? {T}he representation of semantic knowledge in the human brain}.
\bjtitle{Nature Reviews Neuroscience}
\bvolume{8}(\bissue{12}),
\bfpage{976}--\blpage{987}
(\byear{2007})
\end{barticle}
\endbibitem

\bibitem[\protect\citeauthoryear{Kumar}{2021}]{kumar2021semantic}
\begin{barticle}
\bauthor{\bsnm{Kumar}, \binits{A.A.}}:
\batitle{Semantic memory: A review of methods, models, and current challenges}.
\bjtitle{Psychonomic Bulletin \& Review}
\bvolume{28}(\bissue{1}),
\bfpage{40}--\blpage{80}
(\byear{2021})
\end{barticle}
\endbibitem

\bibitem[\protect\citeauthoryear{Jones et~al.}{2015}]{jones2015models}
\begin{barticle}
\bauthor{\bsnm{Jones}, \binits{M.N.}},
\bauthor{\bsnm{Willits}, \binits{J.}},
\bauthor{\bsnm{Dennis}, \binits{S.}},
\bauthor{\bsnm{Jones}, \binits{M.}}:
\batitle{Models of semantic memory}.
\bjtitle{Oxford Handbook of Mathematical and Computational Psychology}
\bvolume{1},
\bfpage{232}--\blpage{54}
(\byear{2015})
\end{barticle}
\endbibitem

\bibitem[\protect\citeauthoryear{Volle}{2018}]{volle2018associative}
\begin{botherref}
\oauthor{\bsnm{Volle}, \binits{E.}}:
Associative and controlled cognition in divergent thinking: Theoretical, experimental, neuroimaging evidence, and new directions.
The Cambridge Handbook of the Neuroscience of Creativity,
333--360
(2018)
\end{botherref}
\endbibitem

\bibitem[\protect\citeauthoryear{Wang et~al.}{2010}]{wang2010idea}
\begin{bchapter}
\bauthor{\bsnm{Wang}, \binits{H.-C.}},
\bauthor{\bsnm{Cosley}, \binits{D.}},
\bauthor{\bsnm{Fussell}, \binits{S.R.}}:
\bctitle{Idea expander: Supporting group brainstorming with conversationally triggered visual thinking stimuli}.
In: \bbtitle{Proceedings of the 2010 ACM Conference on Computer Supported Cooperative Work},
pp. \bfpage{103}--\blpage{106}
(\byear{2010})
\end{bchapter}
\endbibitem

\bibitem[\protect\citeauthoryear{Siangliulue et~al.}{2015}]{siangliulue2015toward}
\begin{bchapter}
\bauthor{\bsnm{Siangliulue}, \binits{P.}},
\bauthor{\bsnm{Arnold}, \binits{K.C.}},
\bauthor{\bsnm{Gajos}, \binits{K.Z.}},
\bauthor{\bsnm{Dow}, \binits{S.P.}}:
\bctitle{Toward collaborative ideation at scale: Leveraging ideas from others to generate more creative and diverse ideas}.
In: \bbtitle{Proceedings of the 18th ACM Conference on Computer Supported Cooperative Work \& Social Computing},
pp. \bfpage{937}--\blpage{945}
(\byear{2015}).
\bcomment{ACM}
\end{bchapter}
\endbibitem

\end{thebibliography}
\end{document}